\documentclass{article}

\PassOptionsToPackage{numbers, compress}{natbib}

\usepackage{natbib}

\usepackage[preprint]{neurips_2023}

\usepackage[utf8]{inputenc} 
\usepackage[T1]{fontenc}    
\usepackage{hyperref}       
\usepackage{url}            
\usepackage{booktabs}       
\usepackage{amsfonts}       
\usepackage{nicefrac}       
\usepackage{microtype}      
\usepackage{xcolor}         
\usepackage{multirow}
\usepackage{bm}

\usepackage{times}
\usepackage{latexsym}

\usepackage{amsmath}

\usepackage{algorithm}
\usepackage{algpseudocode}
\usepackage{graphicx}  
\usepackage{CJK}
\usepackage{bbm}
\usepackage{mathrsfs}
\usepackage{float}
\usepackage{xcolor,colortbl}
\usepackage{MnSymbol}

\usepackage{bigstrut,bigdelim}
\usepackage{paralist}
\usepackage{diagbox}
\usepackage{wrapfig}
\usepackage{verbatim}

\usepackage{subfigure}
\usepackage{multicol}
\usepackage{multirow}
\usepackage{mathtools}

\definecolor{mygreen}{RGB}{2, 70, 3}

\title{Decomposing the Enigma: Subgoal-based Demonstration Learning for Formal Theorem Proving}

\author{%
Xueliang Zhao \\
The University of Hong Kong \\
\texttt{xlzhao22@connect.hku.hk} \\
\And
Wenda Li \\
University of Cambridge \\
\texttt{wl302@cam.ac.uk} \\
\AND
Lingpeng Kong \\
The University of Hong Kong \\
\texttt{lpk@cs.hku.hk} \\
}

\begin{document}

\maketitle

\begin{abstract}
Large language models~(LLMs) present an intriguing avenue of exploration in the domain of formal theorem proving. Nonetheless, the full utilization of these models, particularly in terms of demonstration formatting and organization, remains an underexplored area. In an endeavor to enhance the efficacy of LLMs, we introduce a subgoal-based demonstration learning framework, consisting of two primary elements: Firstly, drawing upon the insights of subgoal learning from the domains of reinforcement learning and robotics, we propose the construction of distinct subgoals for each demonstration example and refine these subgoals in accordance with the pertinent theories of subgoal learning. Secondly, we build upon recent advances in diffusion models to predict the optimal organization, simultaneously addressing two intricate issues that persist within the domain of demonstration organization: subset selection and order determination. Through the integration of subgoal-based learning methodologies, we have successfully increased the prevailing proof accuracy from 38.9\% to 44.3\% on the miniF2F benchmark. Furthermore, the adoption of diffusion models for demonstration organization can lead to an additional enhancement in accuracy to 45.5\%, or a $5\times$ improvement in sampling efficiency compared with the long-standing state-of-the-art method. Our code is available at \url{https://github.com/HKUNLP/subgoal-theorem-prover}.
\end{abstract}

\section{Introduction}

Mathematical theorem proving constitutes a significant milestone in the pursuit of artificial intelligence. Recently, machine learning methodologies have spurred advancements in both formal and informal theorem proving domains~\cite{polu2020generative,lewkowycz2022solving}. Our study falls into the former category. In contrast to informal theorem proving, formal methods have the advantage of leveraging interactive proof assistants~\cite{paulson2000isabelle} to automatically validate proofs generated by models, delegating the verification task to computational systems rather than human intervention. This significantly reduces the costs associated with proof checking, and has been applied in software verification \cite{klein2009sel4} and research-level mathematics \cite{castelvecchi2021mathematicians}. 

Recently, advances in large language models (LLMs) shed new light on the domain of formal theorem proving. The complexity of automated theorem proving comes from the necessity of searching through a vast space of possible logical statements and proof methods, in order to determine the truth-value of a given theorem. LLMs reduce the difficulty of the searching problem by factorizing the formal proof automation task into two in-context learning (\S\ref{ssec:in-context-learning}) problems~\cite{wu2022autoformalization,jiang2022draft,first2023baldur}. Given a mathematical \emph{statement}, an LLM first generates its \emph{informal proof} as a draft. It then generates a \emph{formal sketch} based on this draft, which is ready for an off-the-shelf prover to verify its correctness automatically.\footnote{In practice, the \emph{informal proof} often serves as inline comments in the \emph{formal sketch} to better guide the generation procedure.} In both of these steps, the quality of the demonstrative in-examples either written by humans or generated by machines is the key to the performance of the system.

\begin{figure*}
\centering
\subfigure[{Subgoal-based Proof}] { \label{fig:informal-subgoal-example}
\includegraphics[width=0.55\columnwidth]{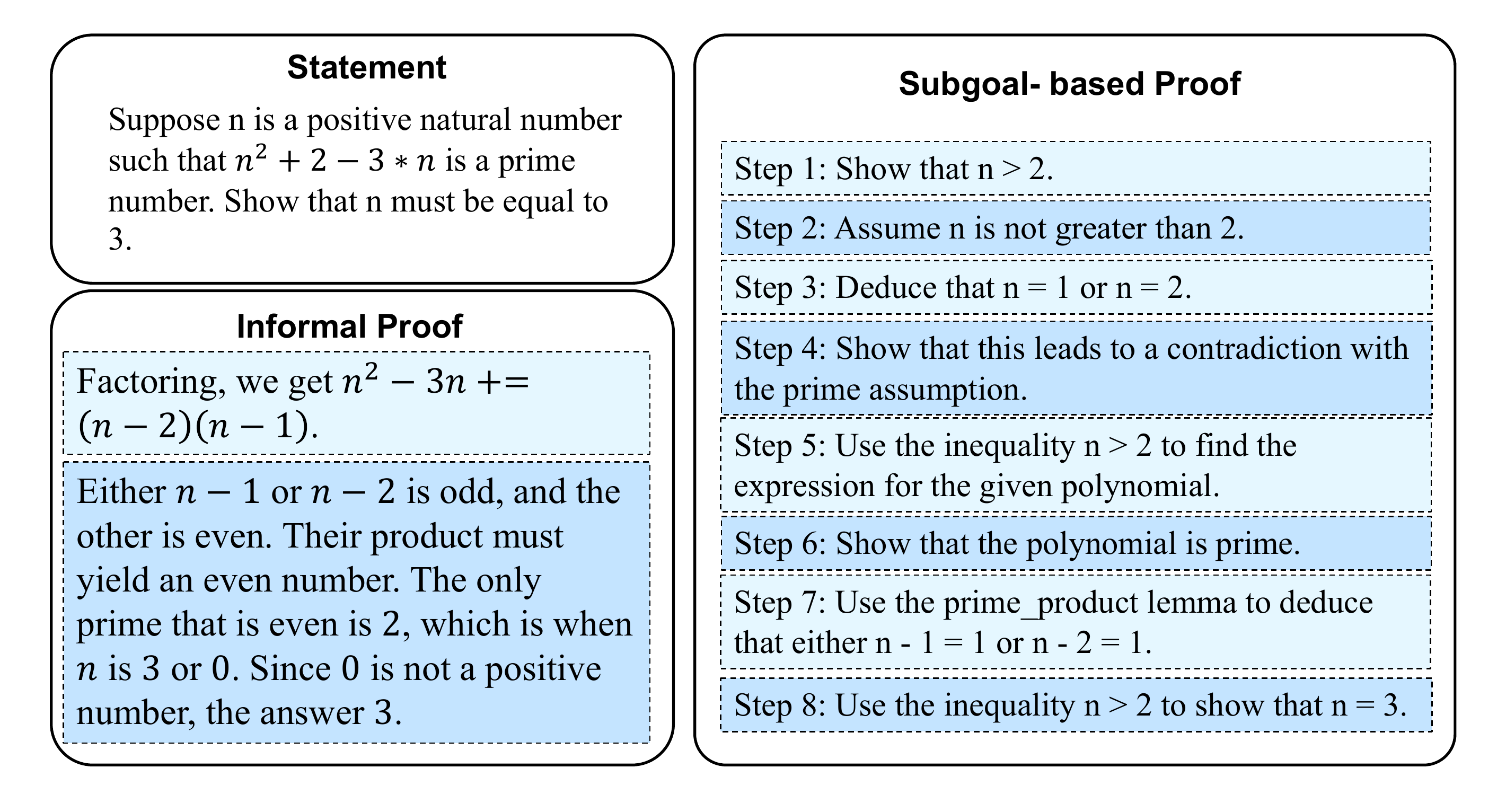}
}
\subfigure[{Demonstration Reorganization}] { \label{diffusion-process-figure1b}
\includegraphics[width=0.35\columnwidth]{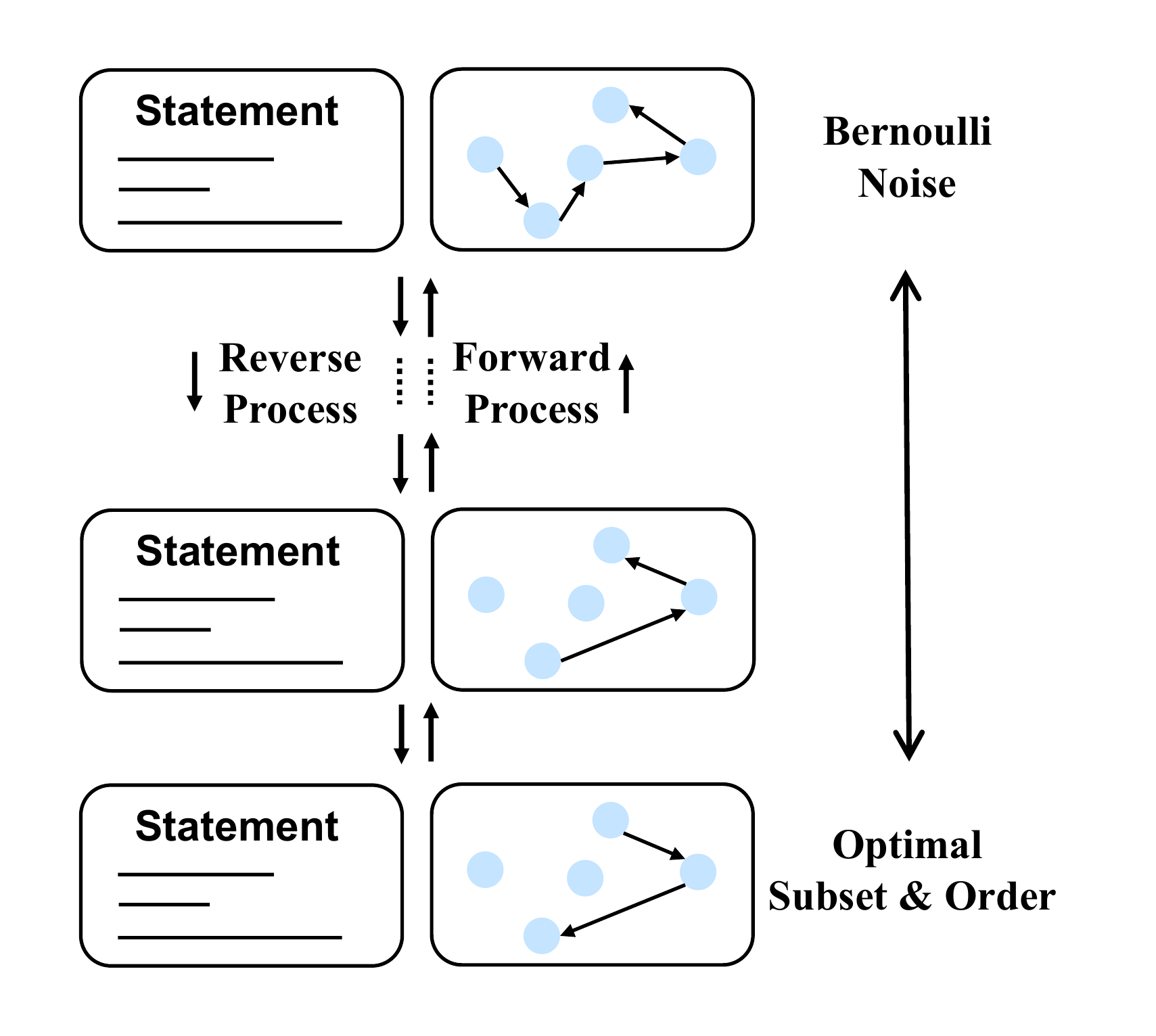}
}
\vspace{-2mm}
\caption{\textbf{Left}: An instance of informal proof and subgoal-based proof. \textbf{Right}: Employing diffusion models to identify a more effective subset of demonstration examples, as well as the optimal order for these examples.}
\end{figure*}

In this paper, we seek to improve the efficacy of LLMs in formal theorem proving by delving deeper into the format and the organization of these demonstrative in-context examples.  We present a subgoal-based demonstration learning framework, comprising two main components. First, we restructure an \emph{informal proof} into a \emph{subgoal-based proof} (Figure~\ref{fig:informal-subgoal-example}), drawing upon the insights of subgoal learning from reinforcement learning and robotics, where studies show that breaking down complex tasks into smaller yet more uniformed subgoals enhances the learning efficiency of the agents\cite{eysenbach2019search,zhang2021c}. 
To construct subgoal-based proofs that can be easily processed and handled by LLMs, we start with human-written informal proofs and then iteratively refine them through interaction with ChatGPT~\cite{team2022chatgpt}, guided by the subgoal learning theory~(\S\ref{sec:subgoal-based formatting}).
Second, a recent study~\cite{wu2022self} points out that the selection and the ordering of the in-context examples have a significant impact on performance. The lengthy formal sketches in automatic theorem proving intensifies this impact, as we can only present very few cases of demonstrations. In response to that, we train a diffusion model to organize the demonstrative in-examples for the translation process from \emph{subgoal-based proof} to its corresponding \emph{formal sketch} of each instance~(\S\ref{sec:diffusion-based organization}).
This approach identifies a more effective subset of demonstration examples as well as the most beneficial order of these examples (Figure~\ref{diffusion-process-figure1b}).

The proposed method significantly outperforms competing approaches in formal theorem proving tasks, achieving a pass rate of $45.5\%$ on miniF2F dataset \cite{zheng2021minif2f}, a $6.6\%$ absolute and $17.0\%$ relative improvement over the previous state-of-the-art system~\cite{jiang2022draft}. Furthermore, the adoption of diffusion models for demonstration organization can lead to a significant improvement in sampling efficiency, reaching previous state-of-the-art ($38.5\%$) on miniF2F with only $20$ (compared to $100$) calls to the LLM.

\section{Subgoal-based Demonstration Learning}
\label{sec:method}
Given a theorem statement $x$, the goal of proof synthesis is to generate a formal sketch $y$ which can be verified by an off-the-shelf automated theorem prover~(e.g., Sledgehammer)~\cite{jiang2022draft}. In this section, we propose the subgoal-based demonstration learning framework which consists of two key components, subgoal-based proof (\S\ref{sec:subgoal-based formatting}) and demonstration reorganization (\S\ref{sec:diffusion-based organization}). The \emph{subgoal-based proof} replaces the \emph{informal proof}, breaking down a complex problem into smaller subgoals that offer more fine-grained and uniform guidance to the LLMs. The \emph{demonstration reorganization} takes place in the stage of generating the \emph{formal sketch} based on the \emph{subgoal-based proof}. This procedure is non-trivial. Given the limited context length of the LLMs, selecting relevant yet diverse demonstration examples has a significant impact on the final pass rate of these formal sketches. 
We denote the set of all $N$ demonstration examples by $E=\{E_1, E_2, \cdots, E_N\}$. Each of them contains a mathematical \emph{statement}, an \emph{informal proof} (or a \emph{subgoal-based proof}), and a \emph{formal sketch}. In the remainder of this section, we first describe the iterative refinement process that produces the subgoal-based proofs given the informal proof, guided by the principles in subgoal learning theory~\cite{zhang2021c}. We then explain our solution to the demonstration reorganization. Starting from collecting arrangements that have yielded successful proofs, we use these as training data for a diffusion model, which progressively determines the most favorable reorganization during inference.

\subsection{Subgoal-based Proof}
\label{sec:subgoal-based formatting}
The significance of LLMs to formal theorem proving is that they grant us the ability to leverage informal proofs to guide formal theorem proving, which otherwise has to be based on expensive heuristics-based brute-force search. Despite considerable progress ~\cite{lewkowycz2022solving,2023arXiv230308774O}, this approach suffers from the flawed informal proofs generated by the LLMs~\cite{jiang2022draft}. We propose to use subgoal-based proofs to replace the informal proofs, where the subgoals are strictly aligned with the states in the automatic provers.  Following \citet{zhang2021c}, we seek to obtain a \emph{valid} sequence of subgoals which satisfies the condition that each subgoal in this sequence should be reachable from both the initial state (i.e., the statement) and the final state (i.e., the passing state of the proof). These valid sequences integrate the guidance from the LLMs better with the search space of the automatic theorem provers, thereby leveraging the ability of the LLMs to the maximum extent. However, it is non-trivial to get these valid subgoal proofs as human-written subgoals often fall short of the above constraints. To address this problem, we iteratively refine the subgoal proof, in the spirit of self-play in reinforcement learning \cite{silver2016mastering}, making calls to both the LLM and the off-the-shelf automated theorem prover.

\paragraph{Subgoal Refinement.}
We start with manually written subgoal-based proofs, and denote these as the initial seed set $\{E^{(0)}_i\}_{i=1}^{N}$. This set contains subgoal-based proofs formed on the informal proofs and the statement, yet not guaranteed to be a valid sequence. We denote the sequence of subgoals in an instance as $(s_0, s_{1}, s_{2}, \cdots, s_{\Delta}, s_{\Delta+1})$, where $\Delta$ is the total number of subgoals and $s_0$ and $s_{\Delta+1}$ are two special subgoals that align with the initial and final states of the automatic prover. During the $k$-th iteration, we randomly select a subset of instances from the previous iteration $\{E^{(k-1)}_i\}_{i=1}^{N}$ as the in-context demonstration for the LLM to generate subgoals for a given instance. According to the definition, $s_{i}$ is considered to be a valid subgoal if and only if it can be reached both from $s_0$ and $s_{\Delta+1}$. Therefore, for each of the subgoal, we recursively call the proof assistant to verify the validness of the most recently developed subgoal and only after $\Delta$ recursions we can obtain the new valid sequence of subgoals and adds that into the next iteration as $E^{(k)}_i$. This process improves the consistency of the derived subgoals in style, thus making it easier for the LLM to learn from in the inference stage. We provide a detailed description of the algorithm in Appendix~\ref{sec:appendix_subgoal}.

\subsection{Demonstration Reorganization}
\label{sec:diffusion-based organization}
The demonstration examples can be lengthy in formal theorem proving. If we assume a maximum context length of $3072$ tokens, only $4.79$ examples on average can be included. Our experiments echo the findings by \citet{wu2022self}. These instance-based demonstration examples have a significant impact on performance. Only certain orders of carefully chosen demonstration examples lead to successful theorem proving. Consequently, identifying the optimal subset from the pool and ordering them into meaningful in-context demonstration examples is of great significance, which unfortunately is an NP-complete problem. We form the demonstration reorganization problem as finding the (Sub)hamiltonian graph where the nodes represent demonstration examples. A traverse following the path corresponds to the selection and the ordering of the in-context examples.
Building upon the recent success of applying diffusion models in addressing NP-complete problems~\cite{graikos2022diffusion,sun2023difusco}, we further formulate this problem into a diffusion process on the graph. This solution has two main advantages. First, it addresses the example selection and ordering problem simultaneously. Second, the inference can be performed in parallel, which greatly reduces the time of discovering the optimal arrangement given the demonstration examples. We start from collecting successful pairs of in-context demonstration example organization and the corresponding statement $x$ as the training data for the diffusion model. We randomly organize (select and order) the demonstration examples and query the LLM to see if it can generate the proof successfully. The passing cases will be used as the starting configuration $\boldsymbol{\psi}_0$ in the diffusion process given the statement $x$.

\paragraph{Training.}
The aim of employing diffusion models is to predict the optimal organization, denoted as $\boldsymbol{\psi}_0$, conditioning on the theorem statement $x$.
From the standpoint of variance inference, diffusion models adopt the following formulations to model $p_{\boldsymbol{\theta}}(\boldsymbol{\psi}_0|x)$,
\begin{equation}
    p_{\boldsymbol{\theta}}(\boldsymbol{\psi}_0|x) \coloneqq \int p_{\boldsymbol{\theta}}(\boldsymbol{\psi}_{0:T}|x) \mathrm{d}\boldsymbol{\psi}_{1:T},
\end{equation}
where $\boldsymbol{\psi}_1, \cdots, \boldsymbol{\psi}_T$ serve as latent variables with the same dimensionality as $\boldsymbol{\psi}_0$.
The learned reverse process progressively denoises these latent variables in order to reconstruct $\boldsymbol{\psi}_0$. This procedure can be formalized as follows,
\begin{equation}
    p_{\boldsymbol{\theta}}(\boldsymbol{\psi}_{0:T}|x)=p(\boldsymbol{\psi}_T)\prod_{t=1}^{T}p_{\boldsymbol{\theta}}(\boldsymbol{\psi}_{t-1}|\boldsymbol{\psi}_t,x).
\end{equation}

The forward process gradually corrupts $\boldsymbol{\psi}_0$ to generate noised latent variables,
\begin{equation}
q(\boldsymbol{\psi}_{1:T}|\boldsymbol{\psi}_0)=\prod_{t=1}^{T}q(\boldsymbol{\psi}_t|\boldsymbol{\psi}_{t-1}).
\end{equation}

The goal of the training process is to maximize the evidence lower bound (ELBO),
\begin{equation}
\begin{aligned}
     \mathbb{E}\left[ \log p_{\boldsymbol{\theta}}(\boldsymbol{\psi}_0|x)\right] \ge 
    \mathbb{E}_q& \left[\log \frac{p_{\boldsymbol{\theta}}(\boldsymbol{\psi}_{0:T}|x)}{q_{\boldsymbol{\theta}}(\boldsymbol{\psi}_{1:T} | \boldsymbol{\psi}_0,x)}\right] \\
    = \mathbb{E}_q  \biggl[\log p_{\boldsymbol{\theta}} (\boldsymbol{\psi}_0 | \boldsymbol{\psi}_1,x) - 
        \sum_{t>1} D_{\mathrm{KL}}[ & q(\boldsymbol{\psi}_{t-1}|\boldsymbol{\psi}_t, \boldsymbol{\psi}_0) \| p_{\boldsymbol{\theta}}(\boldsymbol{\psi}_{t-1}|\boldsymbol{\psi}_t,x)]
    \biggr].
\end{aligned}
\end{equation}
We employ a Graph Neural Network~(GNN) for the encoding and denoising process of the graph. Following \citet{austin2021structured}, we adopt discrete diffusion models to model binary random variables.

\paragraph{Inference.}
During the inference stage, we obtain samples $\boldsymbol{\psi} \sim p_{\boldsymbol{\theta}}(\boldsymbol{\psi}_0 | x)$ and subsequently reconstruct the order of demonstration examples from $\boldsymbol{\psi}$. We then incorporate examples sequentially into the LLM context, and define the output of the demonstration organization module as the sequence of examples upon reaching the LLM length constraint.
More details of the implementation of the diffusion model, the implementation of GNN, and techniques used in the sampling process of $\boldsymbol{\psi}$ can be found in Appendix~\ref{sec:appendix_diff}.

\section{Experiments}
\label{sec:experiment}
\subsection{Formal Environment}
\paragraph{Interactive Theorem Provers.}
Interactive Theorem Provers (ITPs), such as Isabelle~\cite{paulson1994isabelle}, constitute the backbone of contemporary mathematical verification systems. They facilitate the integration of mathematical definitions and theorems into a consistent logical framework, such as Higher-Order Logic or Dependent Type Theory, which is operationalized by their kernels. The kernel plays a pivotal role in the verification process, meticulously examining each theorem to ascertain its recognition by the ITP and thereby ensuring the integrity of the system. The theorem proving process within an ITP is characterized by the articulation of the theorem in the ITP's programming language, followed by an iterative simplification into more manageable objectives or subgoals. The theorem is deemed proven once it can be distilled down to pre-established facts. The selection of Isabelle for our paper is motivated by its intuitive interface, its compatibility with a range of logical frameworks, and its comprehensive library of formalized mathematics.

\paragraph{Sledgehammer.}
Sledgehammer~\cite{paulsson2012three} serves as a powerful tool for automating reasoning within the interactive theorem prover Isabelle. It functions by transmuting the goals encapsulated in Isabelle/HOL's higher-order logic into alternative logics, such as first-order logic. These transmuted goals are then passed to off-the-shelf automated theorem provers, including E, CVC4, Z3, Vampire, and SPASS. In the event that any of these automated theorem provers successfully derives the proof in their respective formats, Sledgehammer undertakes the task of reconstructing the proof within the Isabelle/HOL framework using certified provers, namely metis, meson, and smt. This reconstructed proof, being more interpretable to humans, significantly enhances the system's usability, thereby contributing to the efficiency and effectiveness of (interactive) theorem proving.

\subsection{Dataset and Evaluation}
\label{sec:evaluation}
\paragraph{Dataset.}
We evaluate our approach using the miniF2F dataset~\cite{zheng2021minif2f}, which comprises $488$ formal mathematical problems derived from high-school competitions, expressed in three formal languages: Lean, HOL-Light, and Isabelle. The dataset is divided into a validation and a test set, each including $244$ problems. The problems within the dataset are sourced from three distinct categories: $260$ problems are extracted from the MATH dataset~\cite{hendrycks2021measuring}, $160$ problems are extracted from actual high-school mathematical competitions~(AMC, AIME, and IMO), and $68$ problems are crafted to mirror the difficulty level of the aforementioned competitions.

\paragraph{Evaluation.}
The task at hand entails the generation of formal sketches for problems in the miniF2F dataset. The validity of a formal sketch depends on two criteria: first, the absence of ``cheating'' keywords such as ``sorry'' and ``oops'' that prematurely terminate a proof prior to its completion; second, the capacity of the interactive theorem prover Isabelle to authenticate the corresponding formal statement with the proof. To make working with Isabelle easier, we use the Portal-to-Isabelle API, as introduced by \citet{jiang2022draft}. 
Given the absence of a training split in the miniF2F dataset, we leverage optimal organizations that yield successful proofs from the miniF2F-valid set to train the diffusion model. As proposed by \citet{lample2022hypertree}, we employ the cumulative pass rate as a measure for the results obtained from performing inference using diffusion models on the miniF2F-valid set. This involves integrating the pass rates from both the data collection stage for training and the inference stage.  When it comes to other scenarios, namely conducting inference on the miniF2F-test or cases where the diffusion model is not employed, we simply provide the pass rate.

\subsection{Baselines}
We use the following baselines to test the effectiveness of our proposed methodology.

\paragraph{Symbolic Automated Provers.}
We first employ Sledgehammer, a proof automation tool that is extensively utilized within the Isabelle environment. We adhere to the default configuration of Sledgehammer as provided in Isabelle2021, which encompasses a 120-second timeout and a suite of five automated theorem provers~(Z3, CVC4, SPASS, Vampire, E). In alignment with \citet{jiang2022draft}, we employ Sledgehammer supplemented with heuristics, integrating $11$ prevalent tactics~(i.e., auto, simp, blast, fastforce, force, eval, presburger, sos, arith, linarith, auto simp: field simps) with Sledgehammer.  If all the tactics fail or take longer than $10$ seconds, the system reverts to the base Sledgehammer.

\paragraph{Search-based Methods.}

In addition to the above, we incorporate baselines that utilize Monte-Carlo tree search~\cite{silver2016mastering} to discover the proof. This includes Thor~\cite{jiang2022thor} and another version of Thor that employs an expert iteration on autoformalized data~(i.e., Thor+expert iteration~\cite{wu2022autoformalization}). Thor combines language models with automatic theorem provers to overcome the challenge of selecting beneficial premises from a vast library. Thor+expert iteration enhances a neural theorem prover by training it on theorems that have been automatically formalized.

\paragraph{LLM-based Method.}

Lastly, we incorporate a LLM-based baseline, namely, \emph{Draft, Sketch and Prove}~(DSP)~\cite{jiang2022draft}. DSP turns informal proofs into formal sketches and leverages these formal sketches to steer an automated prover. Notably, we employ the variant of DSP that is implemented with the $540B$ Minerva model~\cite{lewkowycz2022solving}, as this particular implementation demonstrated superior performance in their paper.

We exclude representative methods such as HyperTree Proof Search~(HTPS)~\cite{lample2022hypertree} and GPT-f with expert iteration~\cite{polu2022formal}, which are implemented using Lean~\cite{de2015lean}, a different interactive theorem prover. The disparity in tactics and automation between Lean and Isabelle renders them not directly comparable to our method.

\subsection{Implementation Details}
Throughout our work, we employ ChatGPT~\footnote{the \textit{gpt-3.5-turbo-0301} version} as the LLM. For the creation of the formal sketch, the temperature and max\_tokens
parameters of ChatGPT are set to $0$ and $1024$, respectively. 
In terms of the establishment of the subgoal-based proof, we set the number of refinement iterations to be $15$, with the number of demonstration examples, denoted as $N$, being set to $61$.
For demonstration organization, we employ a randomized demonstration organization approach to generate proofs for $116$ distinct statements on miniF2F-valid, which yield $137$ successful proofs. We then partition the corresponding demonstration contexts into a training set and a validation set, comprising $81$ and $56$ instances respectively. The training of our diffusion models is conducted with a learning rate of $5e-4$, a batch size of $16$, and over a span of $50$ epochs.
We set the number of diffusion steps $T$ to $80$. We employ an early stopping strategy on the validation set and report the performance averaged over three different runs.

\subsection{Main Results}

\begin{table*}[t!]
\centering
\caption{Pass rates on the miniF2F dataset with Isabelle. Numbers in bold denote the best performance. Numbers with a $\star$ correspond to the cumulative pass rate~\cite{lample2022hypertree} since the evaluated statements are part of the training for diffusion models. See \S\ref{sec:evaluation} for more details about cumulative pass rate.}
\label{tab:main_results}
\begin{tabular}{lcc}
\toprule
\multicolumn{1}{c}{\textbf{}} & valid   & test    \\ \midrule
Sledgehammer                  & 9.9\%           & 10.4\%          \\
Sledgehammer+heuristic        & 18.0\%          & 20.9\%          \\
Thor                          & 28.3\%          & 29.9\%          \\
Thor + expert iteration       & 37.3\%          & 35.2\%          \\ 
DSP (540B Minerva)            & 42.6\%          & 38.9\%          \\ \midrule
Ours                          & \textbf{48.0\%}$^{\star}$ & \textbf{45.5\%} \\
\bottomrule
\end{tabular}
\end{table*}

The experiment results, as shown in Table~\ref{tab:main_results}, yield several key observations: (1) Our proposed method achieves a pass rate of 48.0\% on miniF2F-valid and 45.5\% on miniF2F-test, surpassing all competing methods. This superior performance is attributable to the subgoal-based proof coupled with usage of diffusion models for demonstration reorganization; (2) The methods Thor and Thor + expert iteration struggle due to the enormously large action space. This space significantly overshadows that of games, thereby posing challenges to the comprehensive utilization of the Monte Carlo tree search.  Consequently, these methods underperform when compared to LLM-based methods; and (3) DSP has pioneered the introduction of the informal proof, a critical step in the LLM-based formal theorem proving task. However, human-written informal proofs do not offer optimal compatibility with large language models. Our method, grounded in the subgoal-learning theory, is capable of inferring subgoal-based proofs that are more amenable to large language models.

\section{Analysis}
\label{sec:analysis}

\subsection{Ablation Study}
\label{sec:analysis_abl}

\begin{table*}[t!]
\centering
\vspace{-2mm}
\caption{Ablation results on the miniF2F dataset with Isabelle. Numbers with a $\star$ correspond to the cumulative pass rate.}
\label{tab:abl_results}
\begin{tabular}{lcc}
\toprule
\multicolumn{1}{c}{\textbf{}}    & valid & test \\ \midrule
Ours                             & 48.0\%$^{\star}$        & 45.5\%       \\ \midrule
- subgoal \& diffusion & 41.8\%        & 38.5\%       \\
- subgoal               & 44.3\%$^{\star}$        & 40.6\%       \\
- diffusion             & 47.5\%        & 44.3\%      \\ \bottomrule
\end{tabular}
\end{table*}

In our ablation study, we examine four variations of our model on the miniF2F dataset, as detailed in Table \ref{tab:abl_results}. The models include our full method (Ours), and three variants with either the subgoal-based proof, demonstration reorganization, or both components removed.

Our full model achieves the highest performance on the test set. This underscores the importance of both subgoal-based proof and demonstration reorganization. The model without both components showed the lowest performance, further emphasizing the significance of these components. The models missing either the subgoal-based proof or reorganization components also show decreased performance, indicating the substantial role of each component.

\begin{figure}[t!]
\centering
\subfigure[{Subgoal-based Proof}] { \label{fig:analysis_subgoal}
\includegraphics[width=0.45\columnwidth]{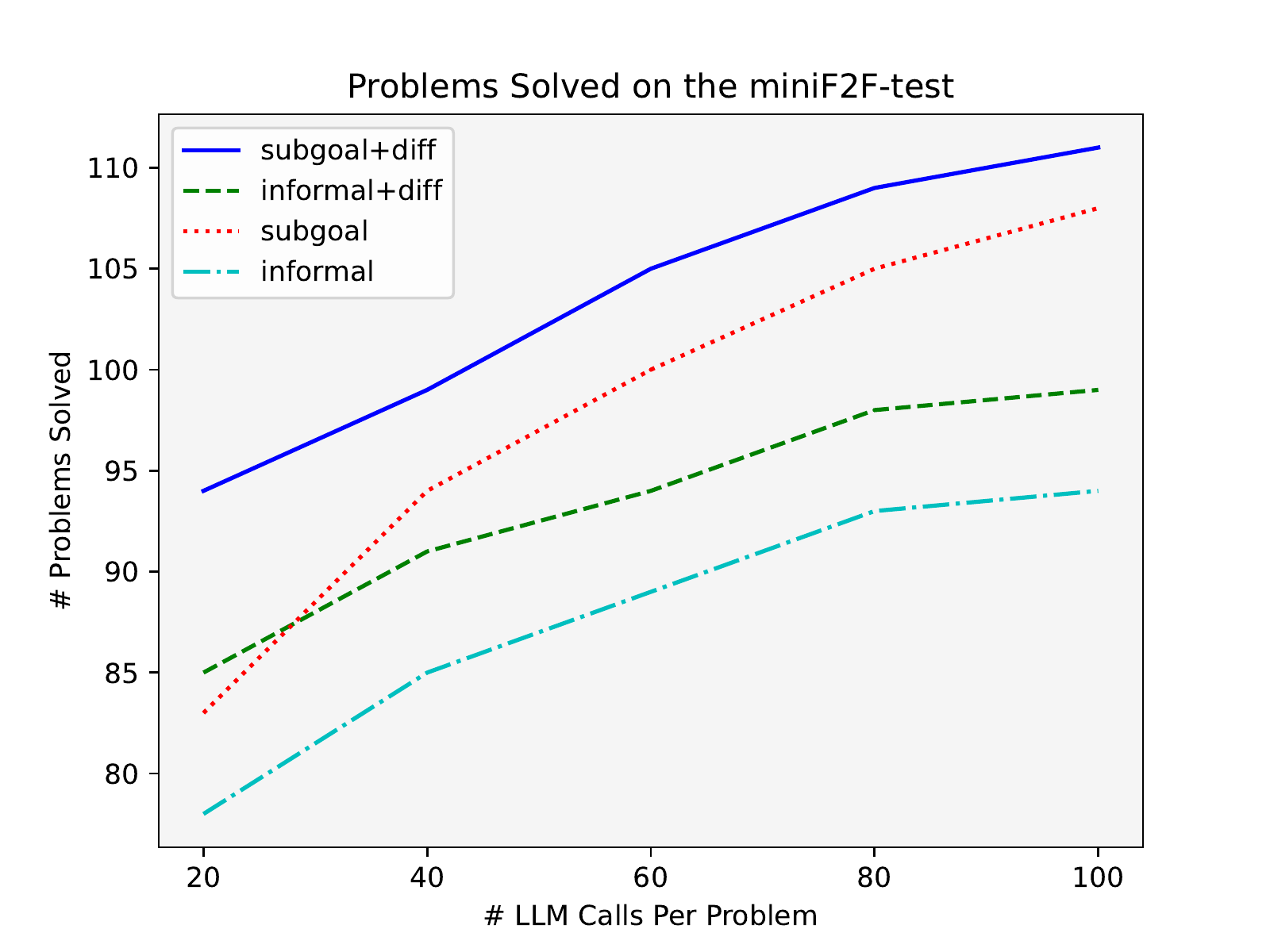}
}
\subfigure[{Demonstration Reorganization}] { \label{fig:analysis_diff}
\includegraphics[width=0.45\columnwidth]{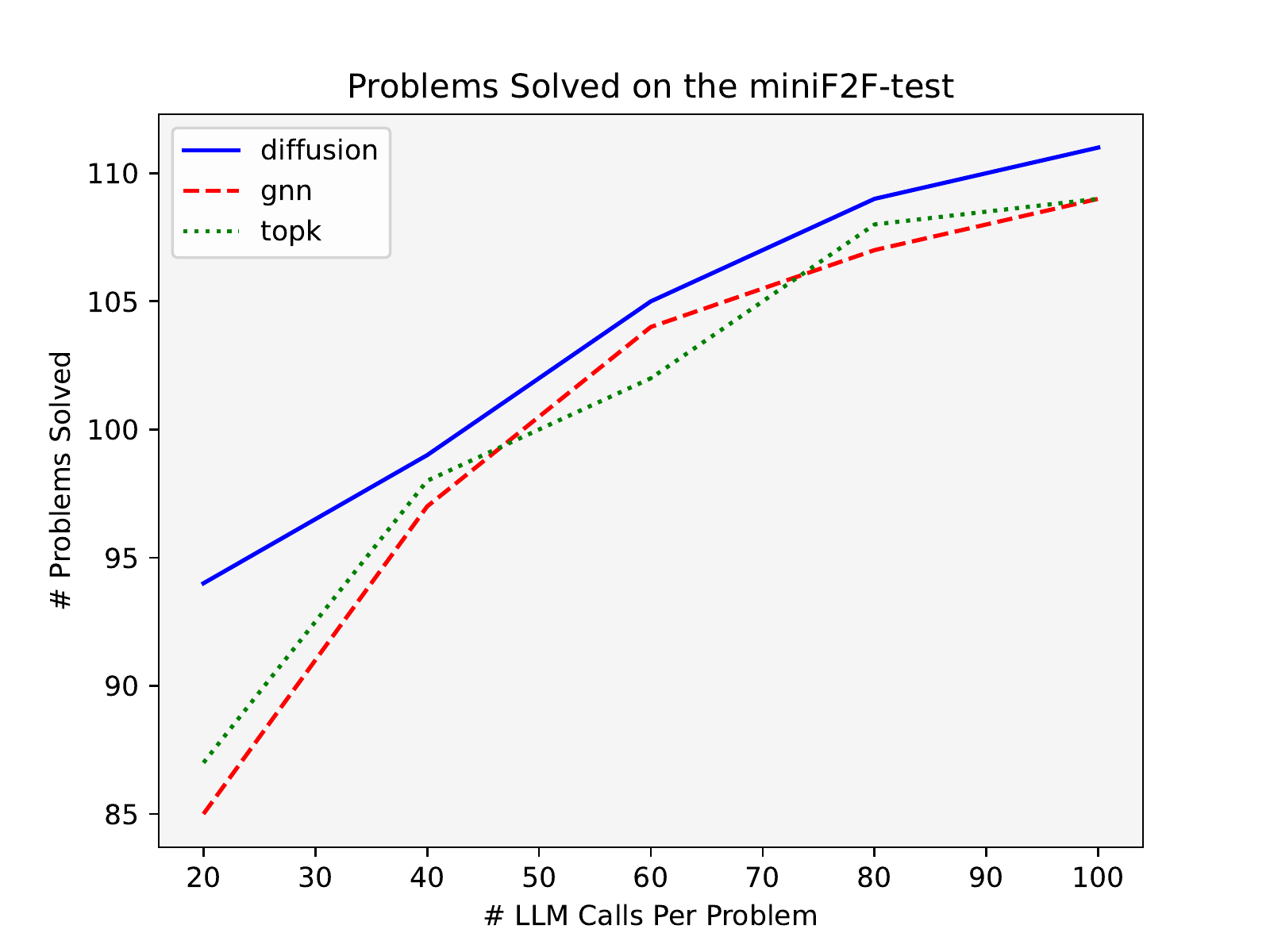}
}
\vspace{-2mm}
\caption{Number of problems solved on miniF2F-test against the number of LLM calls per problem. \textbf{Left}: a comparative assessment between the informal proof and subgoal-based proof under two distinct conditions: presence and absence of the diffusion model. \textbf{Right}: a comparative exploration of different in-context learning methods.}
\end{figure}

\subsection{On the Effect of Subgoal-based Proof}
\label{sec:analysis_subgoal}

We further use four different variants to explore the impact of subgoal-based proof. Figure~\ref{fig:analysis_subgoal} displays the results of this experiment, where ``informal'' denotes the utilization of informal proofs instead of subgoal-based proof, and ``diff'' indicates the integration of demonstration reorganization. The results indicate a significant difference between the approaches that incorporate subgoal-based proof (``subgoal'' and ``subgoal+diff'') and those that do not (``informal'' and ``informal+diff''). This trend remains consistent across all LLM call numbers, suggesting a noteworthy positive effect of subgoal-based proof on the overall performance of our method. 

\subsection{On the Effect of Demonstration Reorganization}
\label{sec:analysis_diff}
To further investigate the effect of a diffusion model for demonstration reorganization, we draw a comparative analysis between its performance and two alternative in-context learning methodologies: the Graph Neural Network (GNN) and the Top-K. The GNN is congruent with a modified version of our proposed model when the inference diffusion step is set to $1$, while the efficacy of the Top-K methodology has been extensively substantiated in the literature~\cite{liu2021makes}.
Figure~\ref{fig:analysis_diff} presents the empirical results, manifesting that the diffusion model's performance increment diminishes as the number of LLM calls escalates to $100$. This phenomenon stems from the fact that the module is trained on data collated from successful proofs via randomized organization sampling. Consequently, it may encounter difficulties in discerning the optimal organization for data that deviates significantly from its training dataset. Nevertheless, this limitation does not overshadow the potential of diffusion models to economize the number of LLM calls.
Notably, with demonstration reorganization, our method exhibits an impressive capability of successfully deriving proofs for $94$ problems~(equivalently, a pass rate of 38.5\%), with a mere $20$ LLM calls. Remarkably, this result is comparable with that of the DSP method, which necessitates $5\times$ the number of LLM calls.

\subsection{Case Study}

\begin{figure*}[t!]
    \centering
    \includegraphics[width=1.0\linewidth]{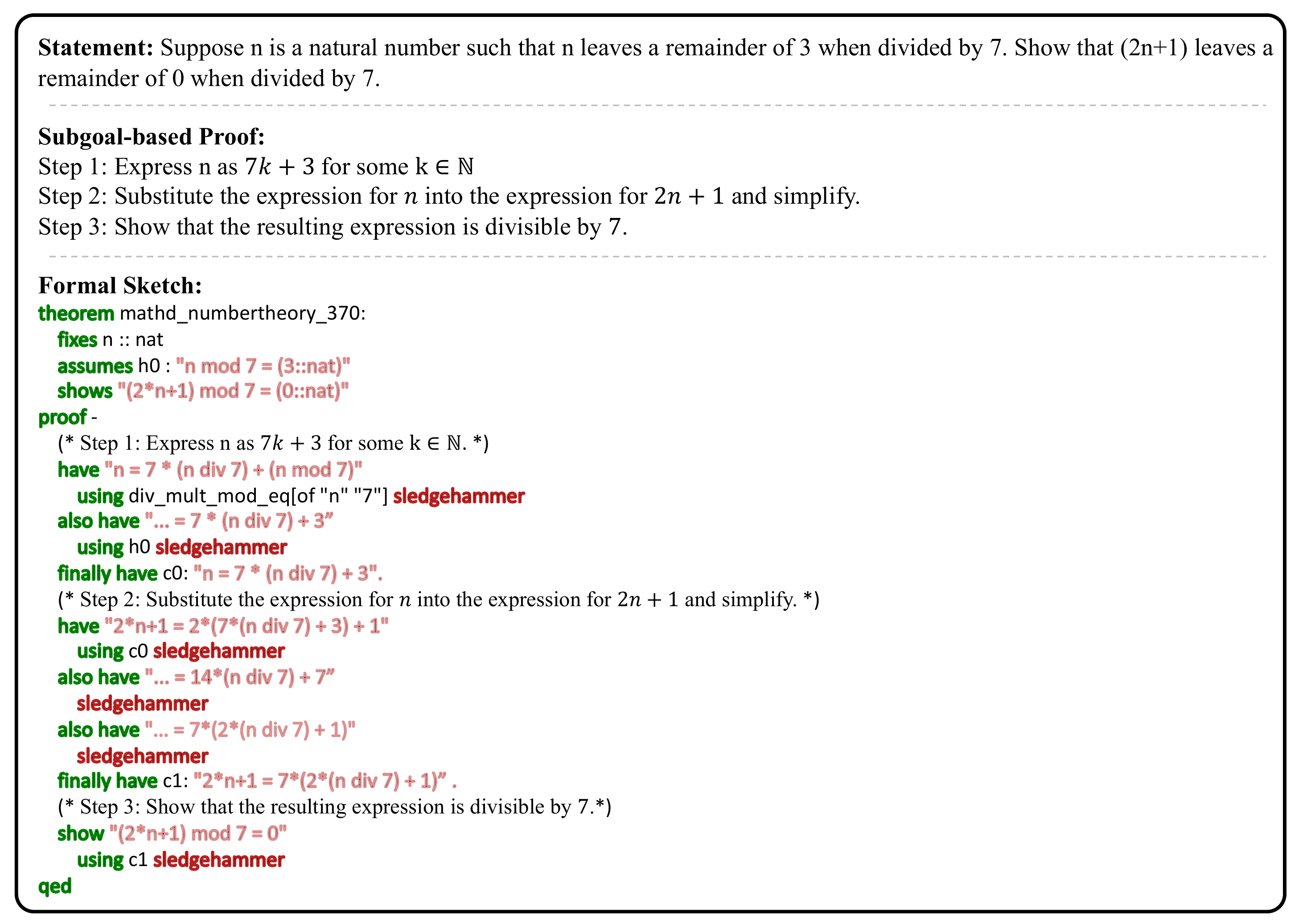}
    \vspace{-5mm}
    \caption{A formal sketch generated by our proposed method.}
    \label{fig:case_study}
    \vspace{-3mm}
\end{figure*}
To better comprehend the efficacy of our proposed method, we present a formal sketch of a problem that remains unproven by earlier state-of-the-art methods.
As demonstrated in Figure~\ref{fig:case_study}, it is apparent that our strategy successfully decomposes the complex objective into three manageable subgoals, each solvable by the LLM. We provide additional comprehensive examples in Appendix~\ref{sec:appendix_case}.

\section{Related Work} 

\subsection{Machine Learning for Formal Theorem Proving}
Machine learning-based formal theorem proving systems primarily fall into two categories: those focusing on proof search strategies and premise selection, and those harnessing Large Language Models (LLMs) for autoformalization and proof generation. The first category, represented by works like Expert Iteration~\cite{polu2022formal} and PACT~\cite{han2021proof}, devise novel learning strategies to enhance proof search, extracting self-supervised data from kernel-level proof terms. Systems such as HyperTree Proof Search (HTPS)\cite{lample2022hypertree} and Thor\cite{jiang2022thor} integrate language models with automated theorem provers, while Magnushammer~\cite{mikula2023magnushammer} presents a transformer-based approach for premise selection. While these techniques have proven effective, they struggle with increasing computational costs as theorems grow more complex. The second category exploits the potential of LLMs in the formalization of mathematical proofs. Both \citet{wu2022autoformalization} and \citet{jiang2022draft} demonstrate that LLMs can convert mathematical problems into formal specifications, with the latter utilizing these translations to guide an automated prover. Baldur~\cite{first2023baldur} extends this approach by generating entire proofs at once and introducing a proof repair model to enhance proving power. However, these approaches have yet to fully leverage the power of LLMs due to a lack of emphasis on the format and organization of demonstration examples. Our work aims to address this gap by introducing a subgoal-based demonstration learning framework that refines the use of LLMs in formal theorem proving.

\subsection{In-context Learning}
\label{ssec:in-context-learning}
In the field of In-Context Learning~(ICL), research primarily focuses on two main areas: (1) the selection of in-context examples, and (2) the arrangement of these examples in the learning context. With regard to the first area, \citet{liu2021makes} suggest a retrieval-based prompt selection method, offering a thoughtful alternative to random example selection. This method aims to find examples that are semantically similar to a test sample to form related prompts. Building on this idea, \citet{rubin2021learning} propose an effective retrieval process for prompts, using a pre-trained language model. \citet{sorensen2022information} further the exploration by introducing a new way to select prompt templates that don't need labeled examples or direct access to the model. Instead, they choose the template that maximizes the mutual information between the input and the corresponding model output. \citet{su2022selective} present a two-step framework that is efficient in annotation. It first selects a set of examples from unlabeled data, and then retrieves task examples from the annotated set during testing. Lastly, \citet{agrawal2022context} focus on creating strategies specifically for machine translation tasks, emphasizing the importance of the quality and domain of in-context examples, and warning against the negative effects of unrelated noisy examples. Works in the second area examine the significance of the order in which prompts are presented. \citet{zhao2021calibrate} point out the instability in few-shot learning caused by the order of training examples and suggest a calibration method to tackle this. \citet{lu2021fantastically} delve deeper into this analysis, demonstrating the sensitivity of prompt order in few-shot learning situations. Even though previous efforts have made remarkable progress in either choosing or sequencing in-context examples, our research sets a new precedent by combining both elements. In this paper, we step out of these isolated areas of concentration, looking into an approach based on diffusion models that effectively tackles both the challenges of selection and ordering at the same time.

\subsection{Subgoal Learning}
Subgoal learning is a pivotal concept in reinforcement learning. It can enable AI systems to solve complex, long-horizon tasks more effectively. Crucially, theoretical analyses have shed light on key concepts including the computational benefits of rewarding subgoals~\cite{zhai2022computational}, the structure of Markov decision processes beneficial for hierarchical reinforcement learning~\cite{wen2020efficiency}, the complexity of optimal option selection for planning~\cite{jinnai2019finding}, and the integration of temporal abstraction into RL~\cite{fruit2017regret}. Empirical analyses in this field mainly focus on subgoal exploration, subgoal generation for planning, and curriculum learning for subgoals. Subgoal exploration aims to find the optimal or efficient exploration of subgoals, employing a variety of strategies. These include minimizing cover time~\cite{jinnai2019discovering}, learning dynamical distances~\cite{hartikainen2019dynamical}, maximizing entropy~\cite{pitis2020maximum}, and utilizing asymmetric self-play~\cite{openai2021asymmetric}. Subgoal planning research encompasses diverse algorithms for improved decision-making. For example, SoRB~\cite{eysenbach2019search} uses RL to build a graph for subgoal sequences, DC-MCTS~\cite{parascandolo2020divide} applies learned sub-goal proposals to partition tasks, PAIR~\cite{li2022phasic} combines online RL and offline supervised learning, and Moro et al.~\cite{moro2022goal} extend MCTS with Hindsight Experience Replay for goal-oriented planning. The research centered on curriculum learning proposes various techniques to create a learning curriculum that gradually intensifies subgoal complexity, thereby optimizing learning efficiency and effectiveness~\cite{zhang2020automatic, zhang2021c}. While there have been preliminary efforts to apply similar principles in the construction of prompts for LLMs~\cite{khot2022decomposed}, the deployment of subgoal learning theories to manage intricate tasks, such as formal theorem proving, remains largely unexplored. Our work pioneers the use of subgoal learning in this domain, with a focus on format and organization.

\section{Conclusion \& Discussion}
In this paper, we have developed a subgoal-based demonstration learning framework that significantly enhances LLMs' efficacy in formal theorem proving. Our approach combines insights from subgoal learning and diffusion models, effectively addressing the challenges of demonstration formatting and organization. As a result, we achieve a 17.0\% relative improvement in proof pass rate on the miniF2F benchmark and a $5\times$ improvement in sampling efficiency. Our work lays the foundation for future endeavors in leveraging AI for generating, validating, and contributing novel insights to automated theorem proving.

Despite the significant advancements achieved through our subgoal-based demonstration learning framework, several limitations of our work exist. Firstly, the process of transforming informal proofs into subgoal-based proofs is an iterative procedure involving interaction with ChatGPT, which may introduce noise and inconsistencies. As our methodology relies on this transformation process, errors introduced at this stage may propagate and affect the final result.
Secondly, while the diffusion models we adopted were effective in organizing the demonstrative in-examples, they are computationally demanding. This can pose challenges for real-time or resource-constrained applications.
Lastly, we only evaluated our framework on the miniF2F dataset. We are expecting to see its performance on other benchmarks and more complex, undergraduate-level mathematical problems~\cite{azerbayev2023proofnet}.

\bibliographystyle{plainnat}
\bibliography{neurips_2023}

\clearpage
\appendix
\section{More Details about Subogal-based Proof}
\label{sec:appendix_subgoal}

We provide a detailed description on the subgoal refinement method~(see §\ref{sec:subgoal-based formatting}) through Algorithm~\ref{alg:subgoal_refinement} and Algorithm~\ref{alg:refine}. In the $k$-th iteration, we construct demonstration examples $\{E_i^{(k)}\}_{i=1}^{N}$ using improved subgoal-based proofs. To construct $E_i^{(k)}$, we first extract the statement and formal sketch from $E_i^{(k-1)}$, then use an LLM to generate subgoals. Afterward, a $\mathrm{Refine}$ module is called to confirm the validity of the created subgoals and adjust any subgoals identified as infeasible.

\begin{figure*}
    \centering
    \includegraphics[width=0.9\linewidth]{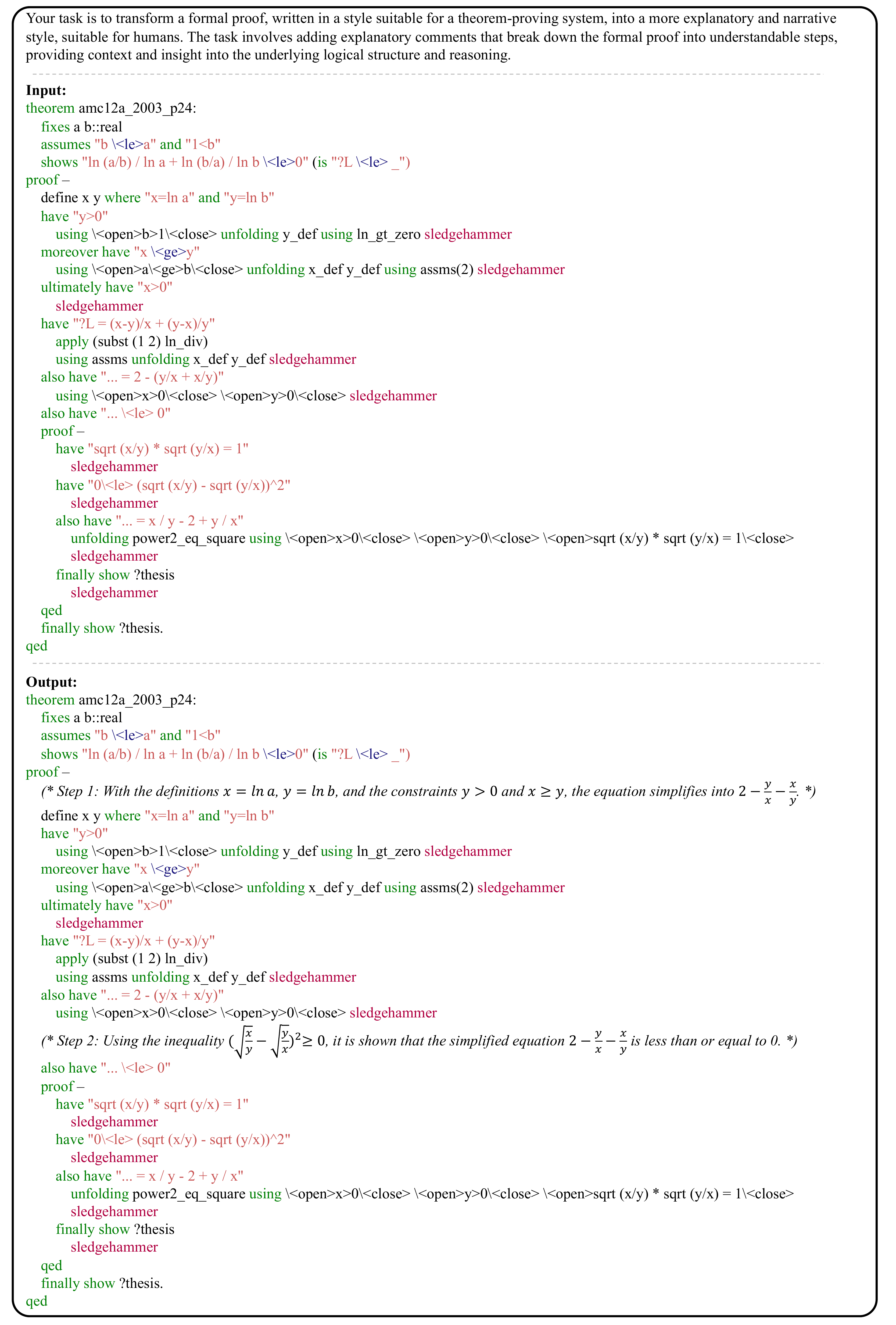}
    \caption{Illustration of the $\mathrm{Initialize\_subgoals}$ function as denoted in Algorithm~\ref{alg:subgoal_refinement}. ChatGPT is leveraged to generate the subgoal-based proof with respect to a formal sketch.}
    \label{fig:initialize_subgoals}
\end{figure*}

\begin{figure*}
    \centering
    \includegraphics[width=0.9\linewidth]{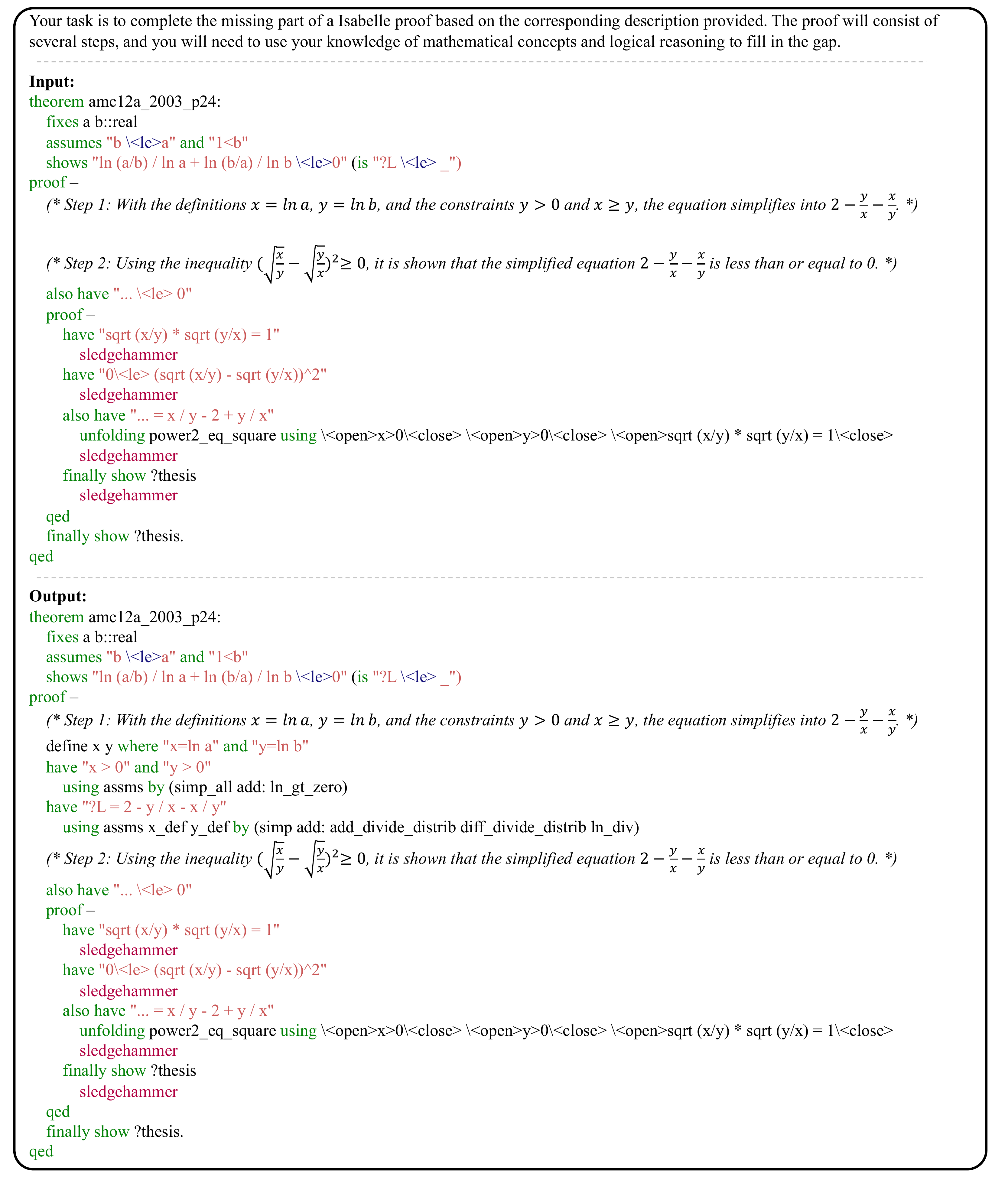}
    \caption{An instance of the ``verify'' component within the $\mathrm{Verify\_and\_correct}$ function in Algorithm~\ref{alg:refine}. ChatGPT encounters a \emph{failure} in reconstructing the proof associated with \emph{step 1}, thereby deeming it an unsuitable subgoal.}
    \label{fig:verify_1}
\end{figure*}

\begin{figure*}
    \centering
    \includegraphics[width=0.9\linewidth]{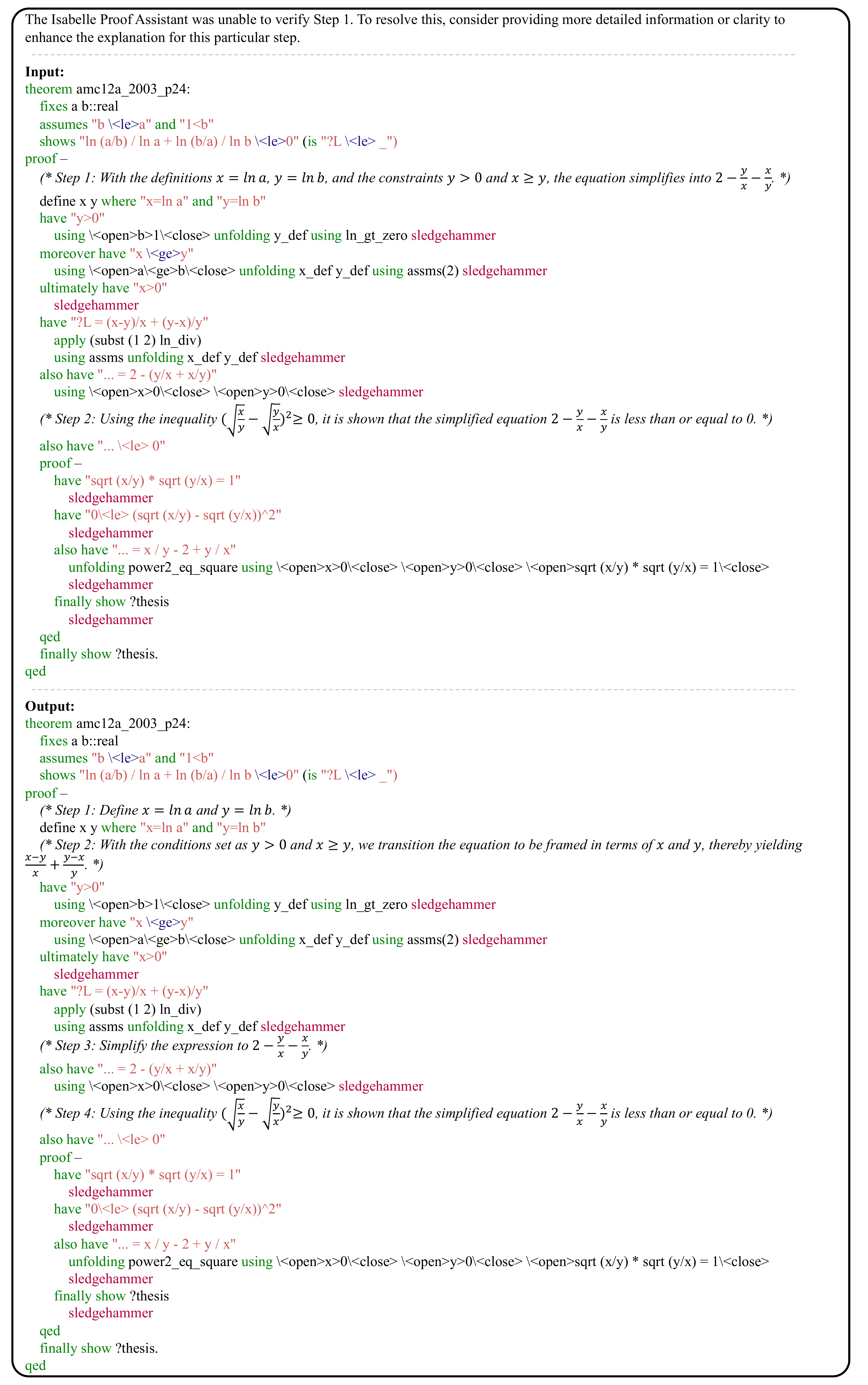}
    \caption{An instance of the ``correct'' component within the $\mathrm{Verify\_and\_correct}$ function in Algorithm~\ref{alg:refine}. ChatGPT works on the decomposition of the original subgoal (i.e., step 1 in the input) into a series of more granular subgoals (i.e., step 1 - 3 in the output).}
    \label{fig:correct_1}
\end{figure*}

\begin{figure*}
    \centering
    \includegraphics[width=0.9\linewidth]{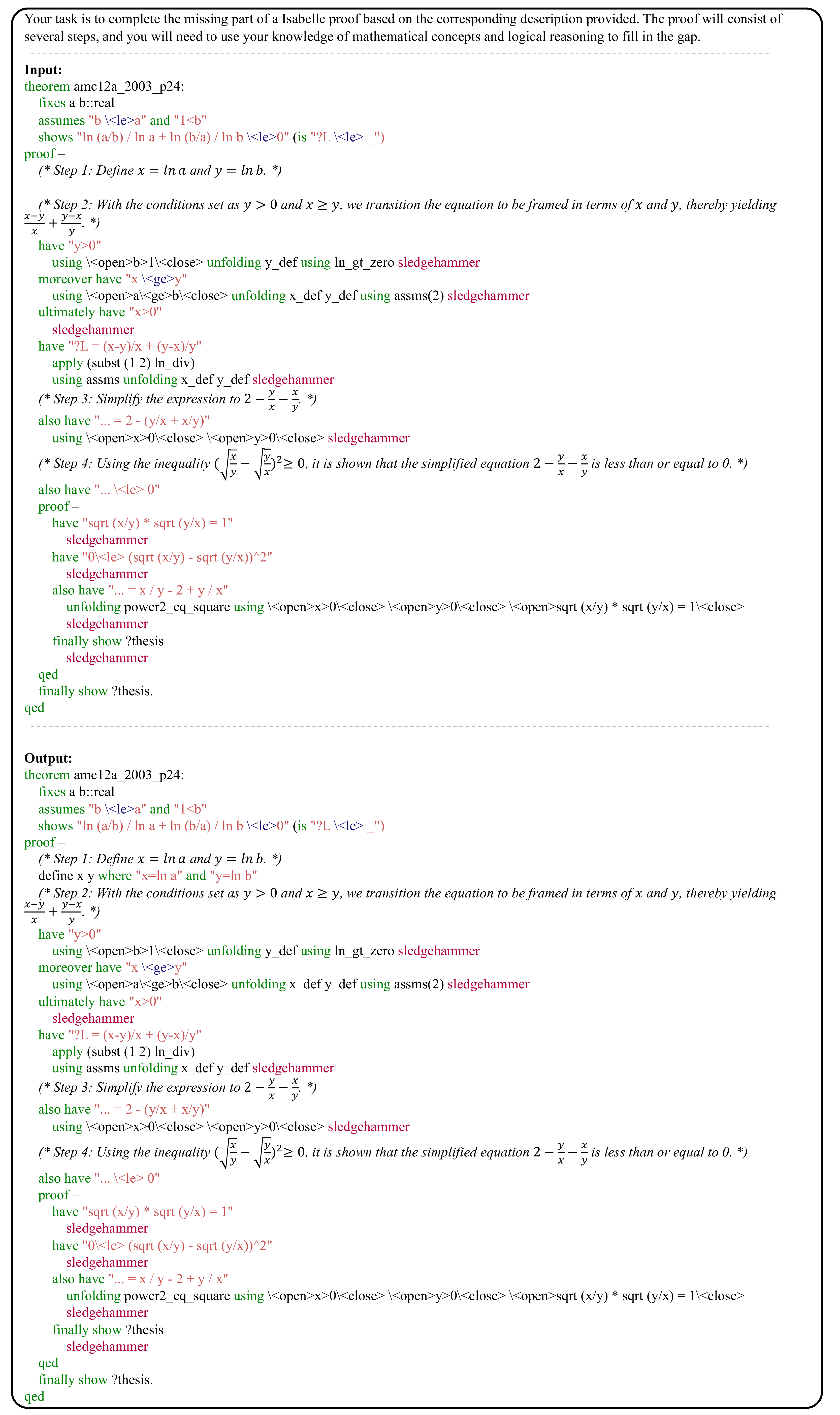}
    \caption{An instance of the ``verify'' component within the $\mathrm{Verify\_and\_correct}$ function in Algorithm~\ref{alg:refine}.  ChatGPT \emph{successfully} reconstructs the proof associated with \emph{step 1}, thus validating it as a viable subgoal.}
    \label{fig:verify_2}
\end{figure*}

\begin{figure*}
    \centering
    \includegraphics[width=0.9\linewidth]{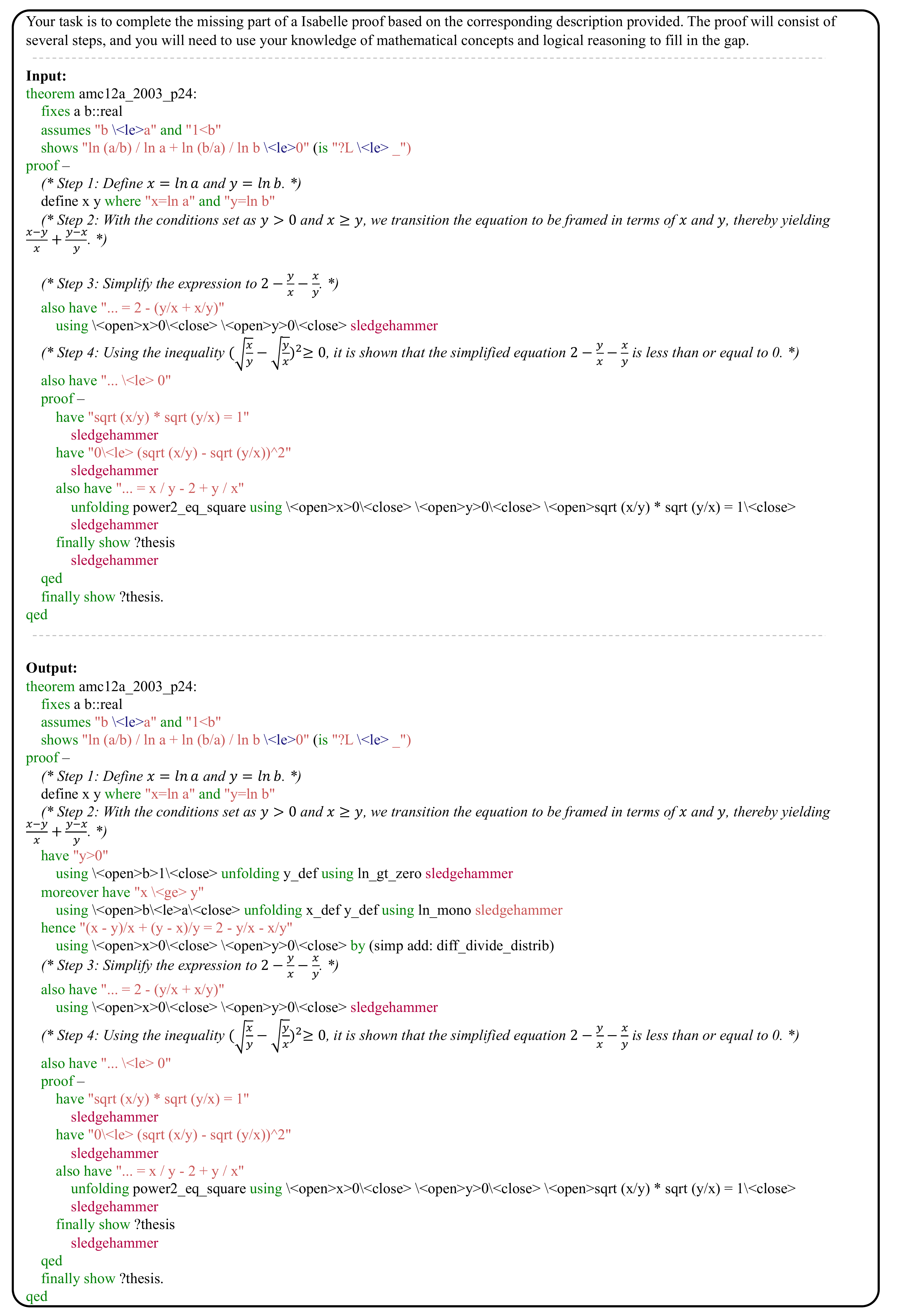}
    \caption{An instance of the ``verify'' component within the $\mathrm{Verify\_and\_correct}$ function in Algorithm~\ref{alg:refine}. ChatGPT encounters a \emph{failure} in reconstructing the proof associated with \emph{step 2}, thereby deeming it an unsuitable subgoal.}
    \label{fig:verify_3}
\end{figure*}

\begin{figure*}
    \centering
    \includegraphics[width=0.9\linewidth]{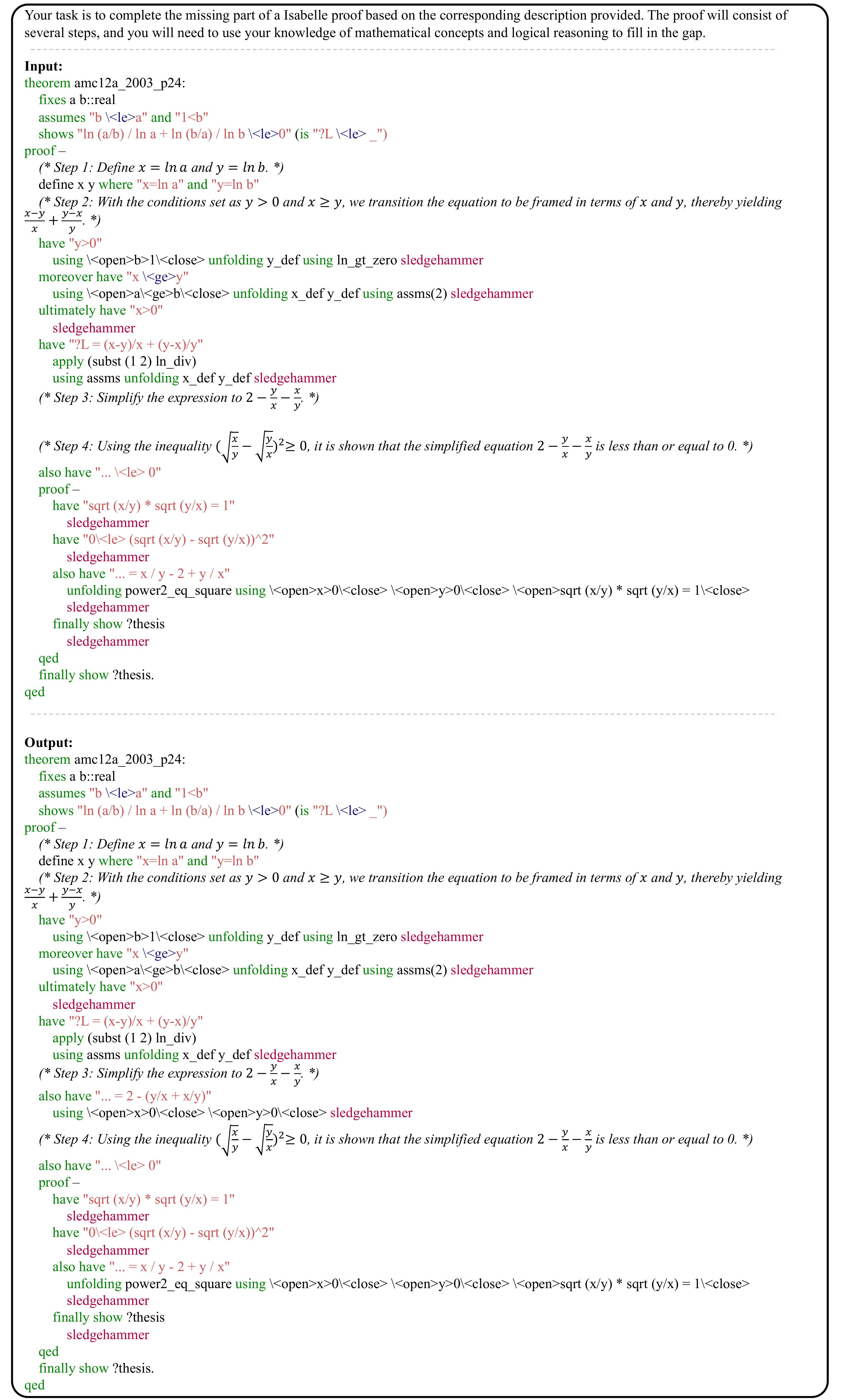}
    \caption{An instance of the ``verify'' component within the $\mathrm{Verify\_and\_correct}$ function in Algorithm~\ref{alg:refine}.  ChatGPT \emph{successfully} reconstructs the proof associated with \emph{step 3}, thus validating it as a viable subgoal.}
    \label{fig:verify_4}
\end{figure*}

\begin{figure*}
    \centering
    \includegraphics[width=0.9\linewidth]{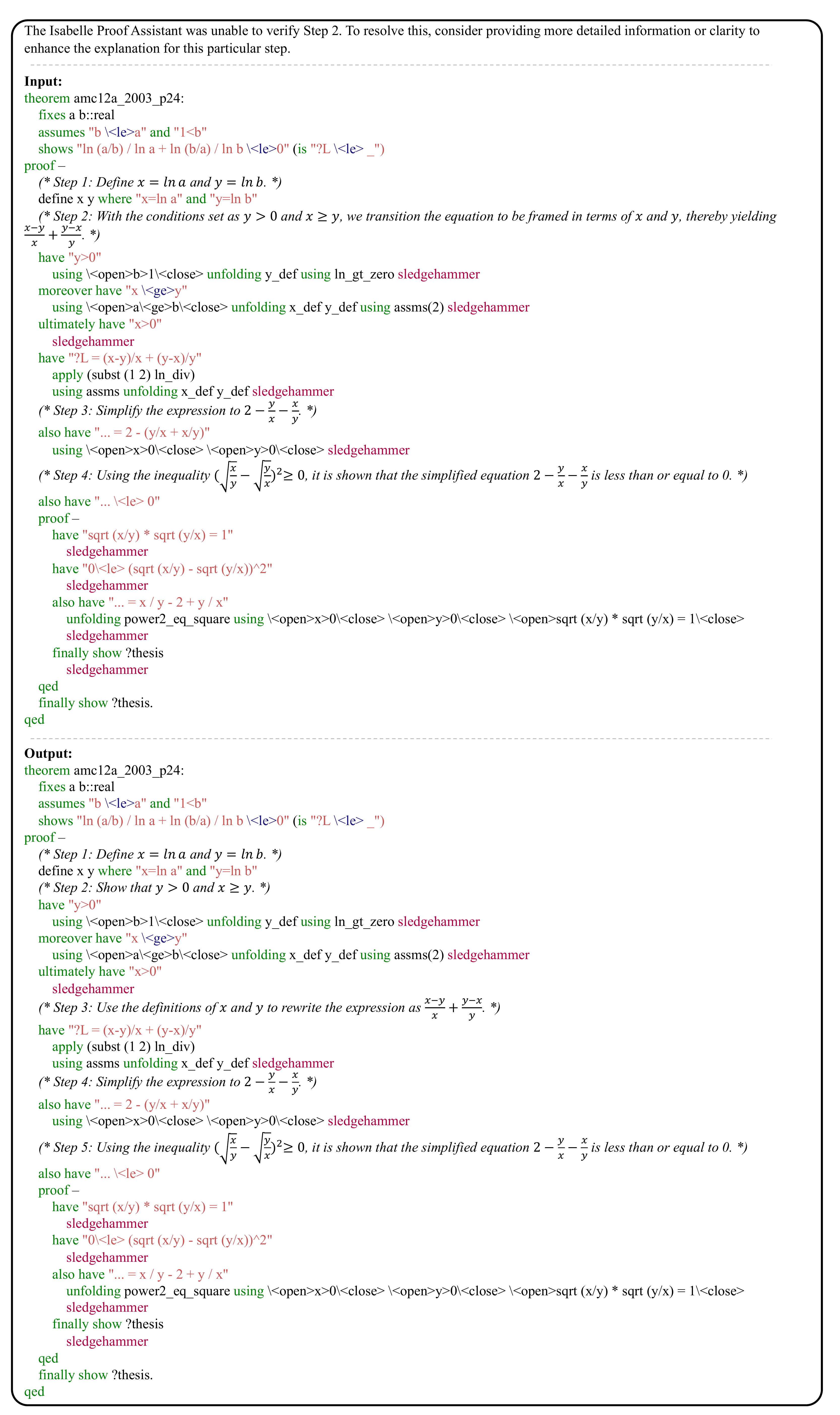}
    \caption{An instance of the ``correct'' component within the $\mathrm{Verify\_and\_correct}$ function in Algorithm~\ref{alg:refine}. ChatGPT works on the decomposition of the original subgoal (specifically, step 2 in the input) into a series of more granular subgoals (specifically, step 2 - 3 in the output).}
    \label{fig:correct_2}
\end{figure*}

\begin{figure*}
    \centering
    \includegraphics[width=0.9\linewidth]{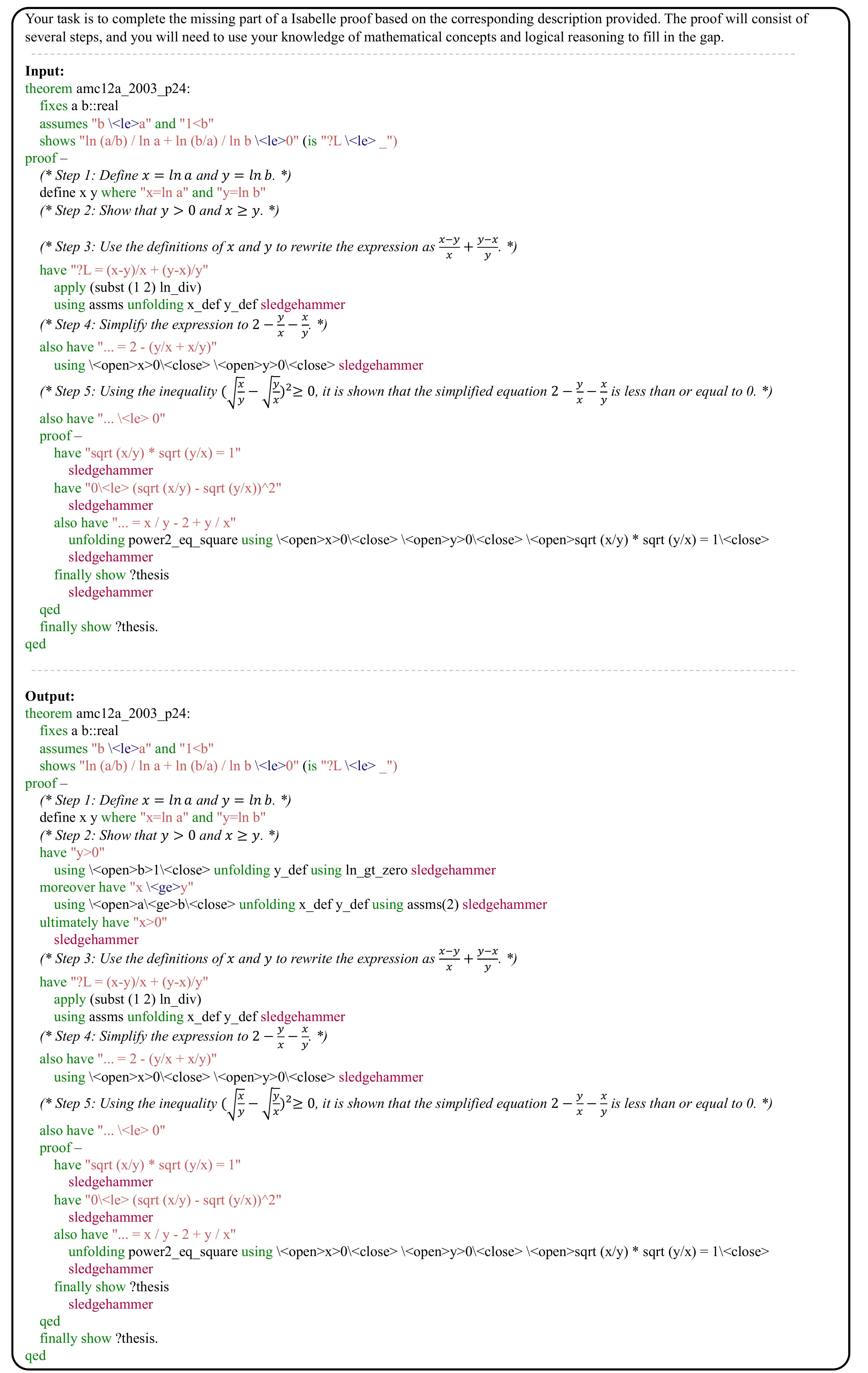}
    \caption{An instance of the ``verify'' component within the $\mathrm{Verify\_and\_correct}$ function in Algorithm~\ref{alg:refine}.  ChatGPT \emph{successfully} reconstructs the proof associated with \emph{step 2}, thus validating it as a viable subgoal.}
    \label{fig:verify_5}
\end{figure*}

\begin{figure*}
    \centering
    \includegraphics[width=0.9\linewidth]{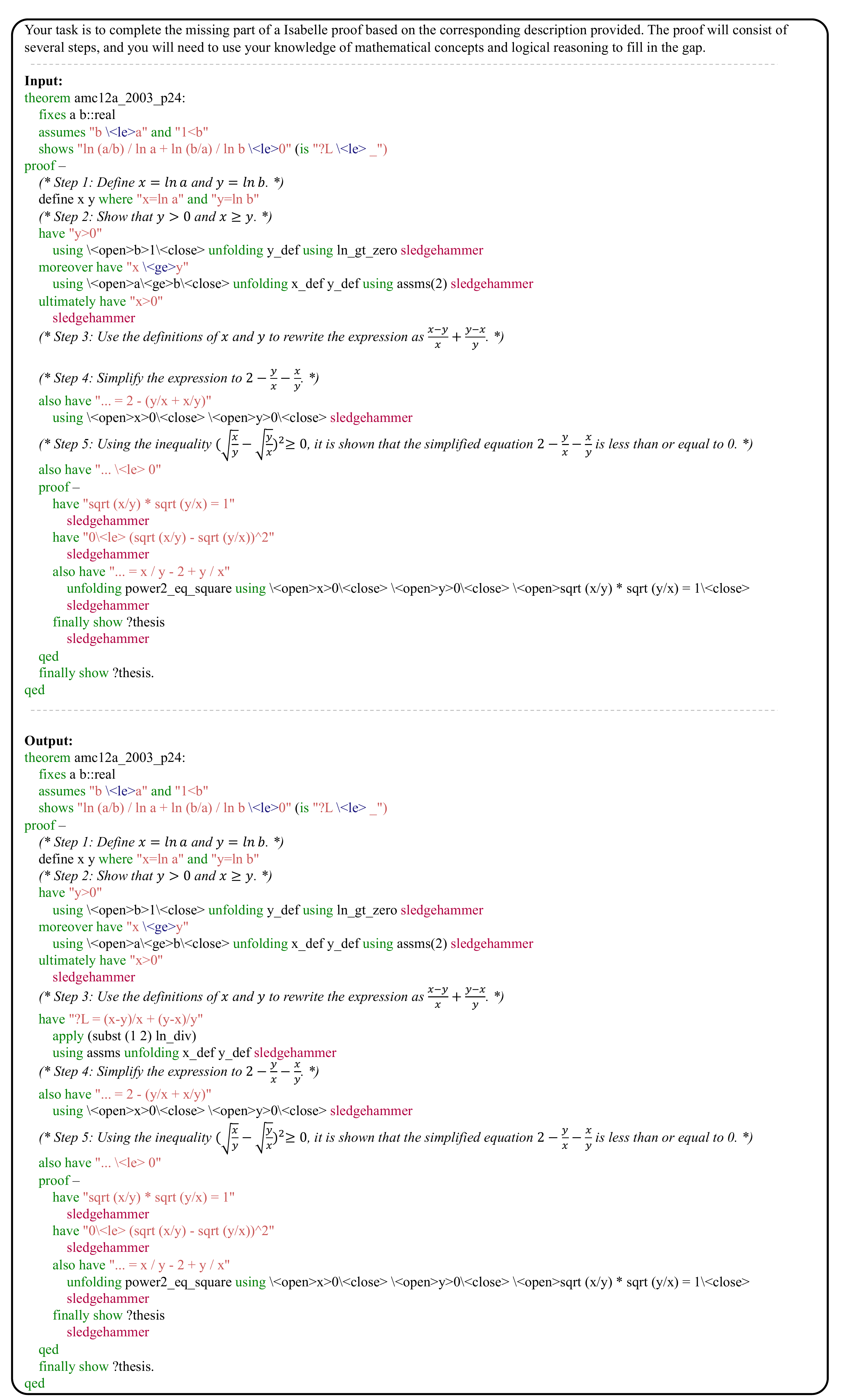}
    \caption{An instance of the ``verify'' component within the $\mathrm{Verify\_and\_correct}$ function in Algorithm~\ref{alg:refine}.  ChatGPT \emph{successfully} reconstructs the proof associated with \emph{step 3}, thus validating it as a viable subgoal.}
    \label{fig:verify_6}
\end{figure*}

We present an example to elucidate this process further~(see Figures~\ref{fig:initialize_subgoals} to~\ref{fig:verify_6}).\footnote{To simplify the illustration, we leave out redundant demonstration examples.} As shown in Figure~\ref{fig:initialize_subgoals}, the LLM creates two subgoals for the theorem \emph{amc12a\_2003\_p4}, leading to $\{s_0, s_1, s_2, s_3\}$.  Refining these subgoals involves calling $\mathrm{verify\_and\_correct}(s_0, s_1)$ to improve the subgoal $s_1$. This is depicted in Figures~\ref{fig:verify_1} to~\ref{fig:verify_6}.  We first use the LLM to reconstruct the subgoal related to the first step, but this attempt fails~(Figure~\ref{fig:verify_1}). Then we break down the subgoal $s_1$ into three more detailed subgoals~(Figure~\ref{fig:correct_1}), each of which is then verified using the same LLM~(Figures~\ref{fig:verify_2} to~\ref{fig:verify_4}). Due to the unsuccessful reconstruction of the second subgoal~(Figure~\ref{fig:verify_3}), it is further broken down into two more specific subgoals~(Figure~\ref{fig:correct_2}). The last two subgoals pass the verification process successfully~(Figures~\ref{fig:verify_5} and~\ref{fig:verify_6}). Finally, the output of $\mathrm{verify\_and\_correct}(s_0, s_1)$, namely $S^{0\rightarrow 1}$, is defined as the set that includes the steps from $1$ to $4$ shown in Figure~\ref{fig:verify_6}.

\begin{algorithm}[]
    \caption{Iterative Subgoal Refinement}
    \label{alg:subgoal_refinement}
\begin{tabular}{ l c l }
    \textbf{Requires: }
    & $\Call{Extract}{}$ & extraction of statement and formal sketch  \\
    & $\Call{Compose}{}$ & composing of a statement, formal sketch \\ 
    & & and subgoals to form a demonstration example \\
    & $\Call{Initialize\_subgoals}{}$ & generate subgoals with a LLM \\ 
\end{tabular}
\begin{algorithmic}
    \Function{iterative\_refinement}{$\{E^{(0)}_1, E^{(0)}_2, \cdots, E^{(0)}_N\}$}
        \For{$k$ in $1,2,\ldots,K$}
            \For{$i$ in $1,2,\ldots,N$}
                \State $x$, $y\gets \Call{Extract}{E^{(k-1)}_{i}}$
                \State $s_0, s_1, \cdots, s_{\Delta}, s_{\Delta+1}\gets \Call{Initialize\_subgoals}{x, y, E^{(k-1)}}$
                \State $S^{0\rightarrow (\Delta+1)}\gets \Call{refine}{(s_0, s_{\Delta+1}, \{s_1, s_2, \cdots, s_{\Delta}\})}$
                \State $E^{(k)}_i\gets \Call{Compose}{x, y, S^{0\rightarrow (\Delta+1)}}$
            \EndFor
        \EndFor
        \State \Return $\{E^{(K)}_1, E^{(K)}_2, \cdots, E^{(K)}_N\}$
    \EndFunction
\end{algorithmic}
\end{algorithm}

\begin{algorithm}[]
    \caption{Refinement Algorithm}
    \label{alg:refine}
\begin{tabular}{ l c l }
    \textbf{Requires: }
    & $\Call{Verify\_and\_correct}{}$ & verify the validness of the subgoals and correct them \\
    & & if necessary \\
\end{tabular}
\begin{algorithmic}
    \Function{refine}{$s_i$, $s_{j+1}$, $\{s_{i+1}, \cdots, s_{j}\}$}
        \If{i = j}
            \Return $\Call{Verify\_and\_correct}{s_i, s_{i+1}}$
        \EndIf
        \State $S^{i\rightarrow i+1}\gets \Call{refine}{s_i, s_{i+1}, \{\}}$
        \State $S^{i+1\rightarrow j+1}\gets \Call{refine}{s_{i+1},s_{j+1},\{s_{i+2}, \cdots, s_{j}\}}$
        \State \Return $S^{i\rightarrow i+1} \cup S^{i+1\rightarrow j+1}$
    \EndFunction
\end{algorithmic}
\end{algorithm}

\clearpage

\clearpage
\section{More Details about Demonstration Reorganization}
\label{sec:appendix_diff}

\subsection{Parameterization}
In alignment with \citet{austin2021structured}, we adopt discrete diffusion models to model binary random variables. Explicitly, the forward process is given by:
\begin{equation}
    q(\boldsymbol{\psi}_{t} | \boldsymbol{\psi}_{t-1}) =
    \mathrm{Cat}\left(\boldsymbol{\psi}_t ; \mathbf{p} = \delta(\boldsymbol{\psi}_{t-1}) \boldsymbol{Q}_t\right),
\end{equation}
where $\delta(\boldsymbol{\psi})$ symbolizes the one-hot encoding of $\boldsymbol{\psi}$, $\boldsymbol{Q}_t=\begin{bmatrix}(1-\beta_t) & \beta_t \\ \beta_t & (1-\beta_t) \end{bmatrix}$ denotes the transition matrix, $\beta_t$ corresponds to the corruption ratio and satisfies that $\prod_{t=1}^{T}(1-\beta_t) \approx 0$.
The marginal at step $t$ and the posterior at step $t-1$ can be articulated as:
\begin{equation}
\begin{aligned}
    q(\boldsymbol{\psi}_{t} | \boldsymbol{\psi}_{0}) &=
    \mathrm{Cat}\left(\boldsymbol{\psi}_t ; \mathbf{p} =  \delta(\boldsymbol{\psi}_{0}) \overline{\boldsymbol{Q}}_t \right), \\
    q(\boldsymbol{\psi}_{t-1}|\boldsymbol{\psi}_t, \boldsymbol{\psi}_0) &= \mathrm{Cat} \left(
\boldsymbol{\psi}_{t-1} ; \mathbf{p} = \frac{
    \delta(\boldsymbol{\psi}_t) \boldsymbol{Q}_t^{\top} \odot \delta(\boldsymbol{\psi}_0) \overline{\boldsymbol{Q}}_{t-1}
}{\delta(\boldsymbol{\psi}_0) \overline{\boldsymbol{Q}}_t \delta(\boldsymbol{\psi}_t)^\top }
\right),
\end{aligned}
\end{equation}
where $\overline{\boldsymbol{Q}}_t=\boldsymbol{Q}_1\boldsymbol{Q}_2\ldots\boldsymbol{Q}_t$. In consonance with \citet{austin2021structured},
we employ a denoising neural network which is tasked with the prediction of $p(\boldsymbol{\psi_0} | \boldsymbol{\psi_t})$, thereby enabling the parameterization of the reverse process:
\begin{equation} 
    p_{\boldsymbol{\theta}}(\boldsymbol{\psi}_{t-1}|\boldsymbol{\psi}_t, x) \propto \sum_{\boldsymbol{\psi}} q(\boldsymbol{\psi}_{t-1}|\boldsymbol{\psi}_t, \boldsymbol{\psi}_0)  p_{\boldsymbol{\theta}}(\boldsymbol{\psi}_0 | \boldsymbol{\psi}_t, x).
\end{equation}

\subsection{Implementation of GNN}

Our work employs a modified version of GNN, a model that exhibits anisotropic characteristics and is enhanced by edge gating methodologies~\cite{bresson2018experimental,sun2023difusco}. We define $\mathbf{t}$ as sinusoidal representations~\cite{vaswani2017attention} associated with the denoising timestep $t$. Consider $\boldsymbol{h}_i^{\ell}$ and $\boldsymbol{e}_{ij}^{\ell}$ as the features of node $i$ and edge $ij$ at a specific layer $\ell$, respectively. During the transition between layers, these features disseminate via an anisotropic message propagation paradigm as follows:
\begin{equation}
\begin{aligned}
\hat{\boldsymbol{e}}^{\ell+1}_{ij} &= \boldsymbol{P}^{\ell} \boldsymbol{e}^{\ell}_{ij} + \boldsymbol{Q}^{\ell} \boldsymbol{h}^{\ell}_i + \boldsymbol{R}^{\ell} \boldsymbol{h}^{\ell}_j, \\
\boldsymbol{e}_{ij}^{\ell+1} &= \boldsymbol{e}_{ij}^{\ell} + \mathrm{MLP}_e(\mathrm{BN}(\hat{\boldsymbol{e}}^{\ell+1}_{ij})) + \mathrm{MLP}_t(\mathbf{t}),\\
\boldsymbol{h}_i^{\ell+1} &= \boldsymbol{h}_i^{\ell} + \mathrm{ReLU}(\mathrm{BN}(\boldsymbol{U}^{\ell} \boldsymbol{h}_i^{\ell}+ \mathrm{SUM}_{j \in \mathcal{N}_i}(\sigma(\hat{\boldsymbol{e}}_{ij}^{\ell+1}) \odot\boldsymbol{V}^{\ell}\boldsymbol{h}^{\ell}_j))),
\end{aligned}
\end{equation}
where $\boldsymbol{P}^{\ell},\boldsymbol{Q}^{\ell},\boldsymbol{R}^{\ell}, \boldsymbol{U}^{\ell},\boldsymbol{V}^{\ell} \in \mathbb{R}^{d\times d}$ denote layer-specific learnable parameters with $d$ denoting the dimension of hidden state.  $\mathrm{BN}$ signifies the Batch Normalization operation~\cite{ioffe2015batch}, while $\mathrm{SUM}$ represents sum pooling. $\odot$ designates the Hadamard product, and $\mathcal{N}_i$ encapsulates the set of neighboring nodes of node $i$. Lastly, a two-layer multi-layer perceptron is denoted by $\mathrm{MLP}{(\cdot)}$.

In our experiments, we define $\boldsymbol{h}_{i}^{0}=\boldsymbol{W} [\mathrm{Ada}(x);\mathrm{Ada}(E^{(K)}_i)]$ where $\boldsymbol{W} \in \mathbb{R}^{d \times 3072}$ is a learnable parameter.  $\mathrm{Ada}(x),\mathrm{Ada}(E^{(K)}_i) \in \mathbb{R}^{1536 \times 1}$ denote the ada embeddings~\footnote{\url{https://platform.openai.com/docs/guides/embeddings}} of the statement $x$ and the $i$-th demonstration example, respectively. The operator $[\cdot;\cdot]$ denotes the concatenation operation between two vectors. $\boldsymbol{e}_{ij}^{0}$ are initialized as sinusoidal features of the edges.

\subsection{Sampling Process}
A straightforward strategy for creating a demonstration organization is by directly sampling $\boldsymbol{\psi} \sim p_{\boldsymbol{\theta}}(\boldsymbol{\psi}_0|x)$. However, this strategy introduces two key challenges: (1) A cycle in $\boldsymbol{\psi}$ might be present, indicating that at least one demonstration example is selected multiple times; (2) $\boldsymbol{\psi}$ could include multiple separate sub-graphs, making it difficult to define the relative position between two demonstration examples from two different sub-graphs. Taking a cue from treating diffusion models as discriminative approaches~\cite{li2023your}, we start by randomly creating $200$ potential solutions. Using the diffusion model's ability to provide conditional density estimates, we rate these $200$ potential solutions and select the one with the highest score to build the final demonstration organization. We then reconstruct the sequence of demonstration examples from $\boldsymbol{\psi}$, adding examples one by one into the LLM context until we hit the length limit of the LLM.

\subsection{Hyperparameters and Hardware Setup}

In the course of our experiment, we employ a $3$-layer Anisotropic Graph Neural Network with a hidden state dimensionality set to $256$.  We sweep the learning rate from $[1e-4, 2e-4, 5e-4, 7e-4]$ and sweep batch size from $[4, 8, 16, 32]$. The processes of training and inference for the diffusion models are conducted on a NVIDIA RTX 3090 GPU.

\section{Additional Examples}
\label{sec:appendix_case}
\begin{figure*}
    \centering
    \vspace{-5mm}
    \includegraphics[width=0.9\linewidth]{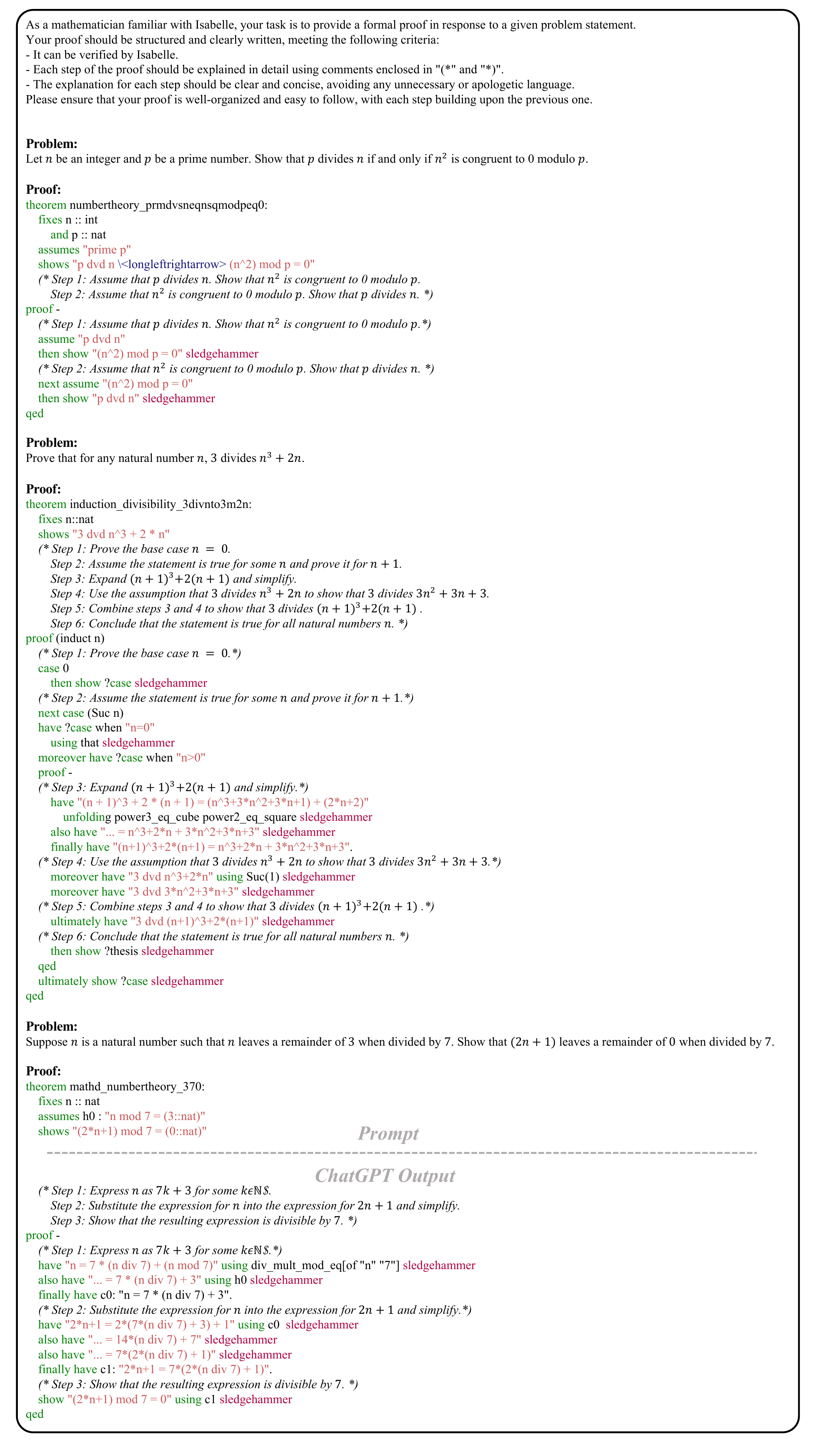}
    \vspace{-3mm}
    \caption{A case from miniF2F-valid. Our method identifies viable subgoals and leverages demonstration examples with division and modulus operations to complete the proof.}
    \label{fig:case_1}
\end{figure*}

\begin{figure*}
    \centering
    \vspace{-14mm}
    \includegraphics[width=0.9\linewidth]{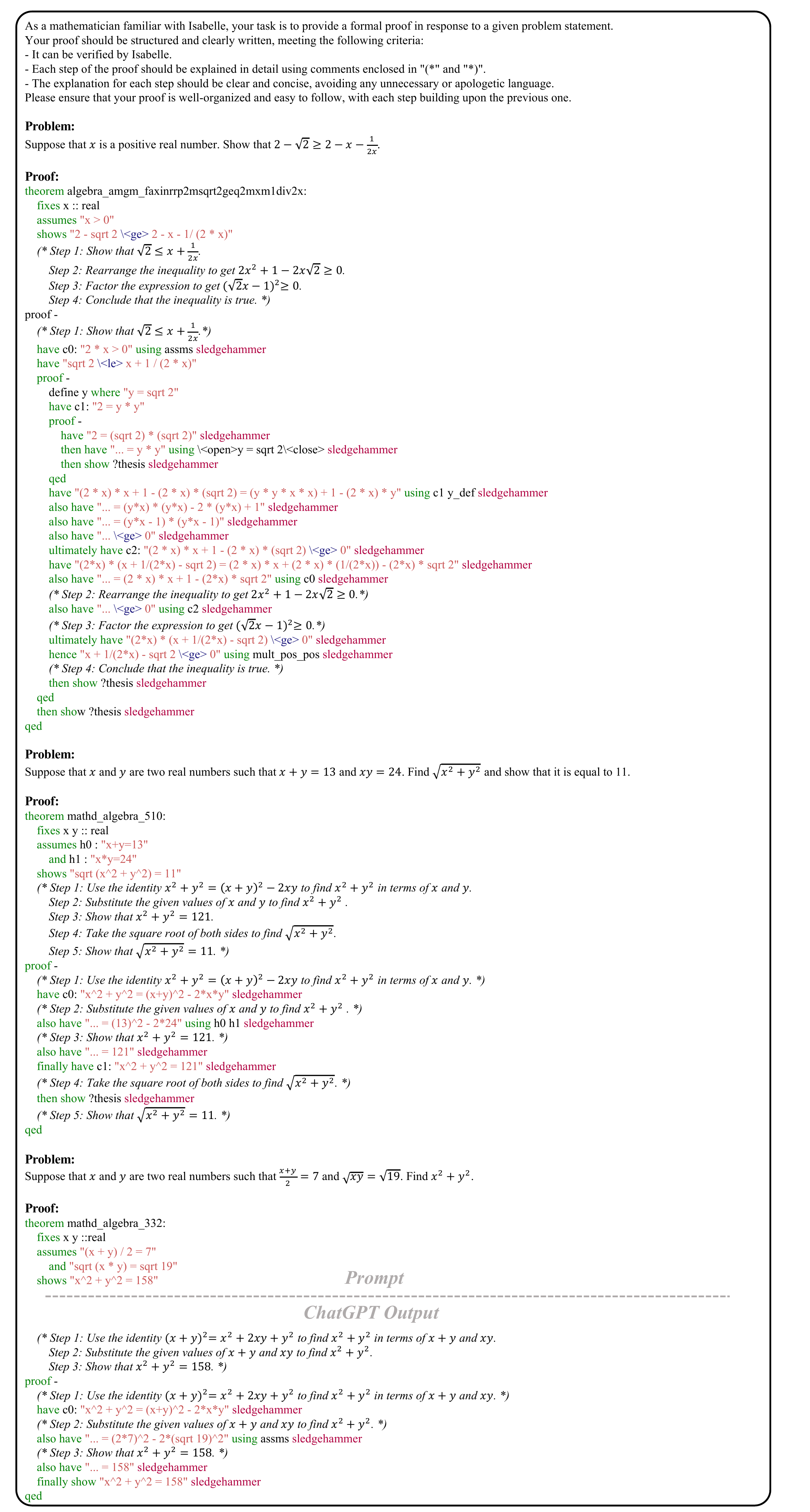}
    \vspace{-3mm}
    \caption{A case from miniF2F-test. Our method predicts viable subgoals and utilizes demonstration examples involving squaring and square root operations to finalize the proof.}
    \label{fig:case_2}
\end{figure*}

\begin{figure*}
    \centering
    \vspace{-10mm}
    \includegraphics[width=0.9\linewidth]{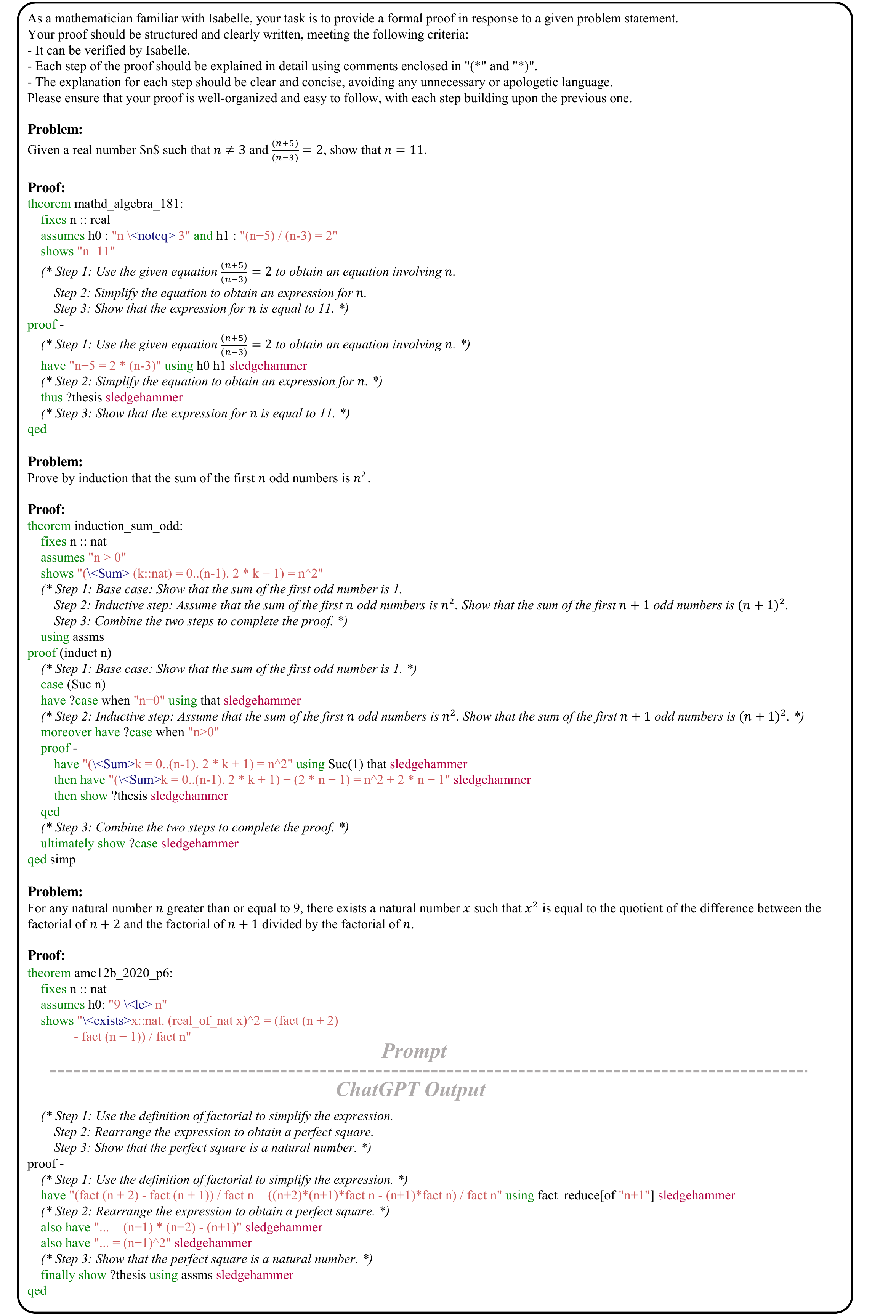}
    \vspace{-3mm}
    \caption{A case from miniF2F-test. Our method recognizes viable subgoals and successfully finishes the proof.}
    \label{fig:case_3}
\end{figure*}

We provide additional cases in Figure~\ref{fig:case_1} to Figure~\ref{fig:case_3}. 
In Figure~\ref{fig:case_1}, our method proficiently identifies viable subgoals, successfully guiding a clear path to the proof. This is accomplished by leveraging pertinent demonstration examples which utilize operations of division and modulus. Akin to the previous case, in Figure~\ref{fig:case_2}, our method accurately predicts potential subgoals. This is realized by capitalizing on demonstration examples that employ the mathematical operations of squaring and square root. Finally, in Figure~\ref{fig:case_3}, our method demonstrates its capacity to consistently discern viable subgoals, thereby achieving a seamless completion of the proof structure. 

\end{document}